\documentclass[10pt,journal,compsoc]{IEEEtran}

\ifCLASSOPTIONcompsoc
  \usepackage[nocompress]{cite}
\else
  \usepackage{cite}
\fi

\usepackage{graphicx}

\usepackage{tikz}
\usepackage{comment}
\usepackage{amsmath,amssymb} 
\usepackage[colorlinks,linkcolor=red]{hyperref}

\usepackage{multirow}
\usepackage{algpseudocode}  
\usepackage{amsmath}  
\usepackage{array}
\usepackage[ruled,linesnumbered]{algorithm2e}
\usepackage{hyperref}
\usepackage{colortbl}

\usepackage{diagbox}

\usepackage{booktabs}
\usepackage{amsmath}  
\usepackage{amssymb}  

\usepackage{amsmath}
\DeclareMathOperator*{\argmin}{arg\,min}  

\begin{document}
%
\title{Real-world Adversarial Defense against Patch Attacks based on Diffusion Model}

\author{
Xingxing~Wei,~
Caixin~Kang,~
Yinpeng~Dong,~
Zhengyi~Wang,~
Shouwei~Ruan,~
Yubo~Chen,~
and~Hang~Su
\IEEEcompsocitemizethanks{\IEEEcompsocthanksitem Xingxing Wei, Caixin Kang, Shouwei Ruan, and Yubo Chen are with the Institute of Artificial Intelligence, Beihang University, No.37, Xueyuan Road, Haidian District, Beijing, 100191, P.R. China. (E-mail: {xxwei, caixinkang}@buaa.edu.cn).

Yinpeng Dong, Zhengyi Wang, and Hang Su are with the Institute for Artificial Intelligence, Beijing National Research Center for Information Science and Technology, Department of Computer Science and Technology, Tsinghua University, Beijing 100084, China. 
}
}

%
%

\markboth{}%
{Shell \MakeLowercase{\textit{et al.}}: Bare Demo of IEEEtran.cls for Computer Society Journals}

\IEEEtitleabstractindextext{%
\begin{abstract}

Adversarial patches present significant challenges to the robustness of deep learning models, making the development of effective defenses become critical for real-world applications. This paper introduces \textbf{DIFFender}, a novel \textbf{DIF}fusion-based De\textbf{Fender} framework that leverages the power of a text-guided diffusion model to counter adversarial patch attacks. At the core of our approach is the discovery of the Adversarial Anomaly Perception (AAP) phenomenon, which enables the diffusion model to accurately detect and locate adversarial patches by analyzing distributional anomalies. DIFFender seamlessly integrates the tasks of patch localization and restoration within a unified diffusion model framework, enhancing defense efficacy through their close interaction. Additionally, DIFFender employs  an efficient few-shot prompt-tuning algorithm, facilitating the adaptation of the pre-trained diffusion model to defense tasks without the need for extensive retraining. Our comprehensive evaluation, covering image classification and face recognition tasks, as well as real-world scenarios, demonstrates DIFFender’s robust performance against adversarial attacks. The framework’s versatility and generalizability across various settings, classifiers, and attack methodologies mark a significant advancement in adversarial patch defense strategies.
Except for the popular visible domain, we have identified another advantage of DIFFender: its capability to easily expand into the infrared domain. Consequently, we demonstrate the good flexibility  of DIFFender, which can defend against both infrared and visible adversarial patch attacks alternatively using a universal defense framework.
\end{abstract}

}

\maketitle

\IEEEdisplaynontitleabstractindextext

%
\IEEEpeerreviewmaketitle

\IEEEraisesectionheading{\section{Introduction}\label{sec:introduction}}

%
%
%
%

\IEEEPARstart{D}{eep} neural networks are susceptible to adversarial examples \cite{szegedy2013intriguing,goodfellow2014explaining}, where small, often imperceptible perturbations are deliberately introduced to natural images, causing the model to make erroneous predictions with high confidence. The majority of adversarial attacks and defenses have focused on $\ell_p$-norm threat models \cite{goodfellow2014explaining,carlini2017towards,dong2018boosting,madry2017towards}, which constrain adversarial perturbations within an $\ell_p$-norm boundary to ensure they remain imperceptible. However, these conventional $\ell_p$-based perturbations necessitate altering every pixel of an image, a method that is typically impractical in physical environments. In contrast, adversarial patch attacks \cite{brown2017adversarial,karmon2018lavan,li2021generative,wei2022adversarial}, which focus perturbations on a specific region of the object, are more feasible in real-world scenarios. These patch-based attacks pose substantial threats to applications such as face recognition \cite{Sharif2016Accessorize,xiao2021improving} and autonomous driving \cite{jing2021too,dong2023benchmarking}.

Despite the numerous adversarial defenses against patch attacks proposed in recent years, their performance remains insufficient to ensure the safety and reliability required for real-world applications. Some approaches rely on adversarial training \cite{wu2019defending, rao2020adversarial} and certified defenses \cite{gowal2019scalable, chiang2020certified}, which tend to be effective only against specific types of attacks and often fail to generalize well to other forms of patch attacks in practical scenarios \cite{nie2022diffusion}. Another category of defenses involves pre-processing techniques \cite{hayes2018visible,naseer2019local,yu2021defending,liu2022segment}, which aim to neutralize adversarial patches through methods like image completion or smoothing. However, these techniques frequently struggle to preserve the high fidelity of the original images, resulting in visual artifacts in the reconstructed images that can negatively affect recognition performance. Moreover, these defenses are vulnerable to stronger adaptive attacks that exploit gradient obfuscation \cite{athalye2018obfuscated}, further limiting their effectiveness.

\begin{figure*}
  \centering
  \includegraphics[width=0.95\linewidth]{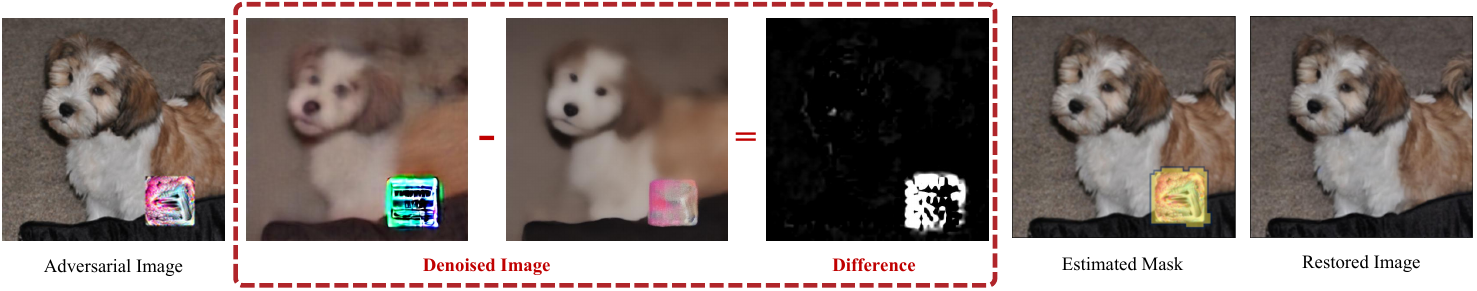}
  \caption{\small 
  The intriguing phenomenon of the diffusion model. When applied multiple times to an adversarial image, the differences between any two resulting denoised images are particularly pronounced within the regions containing adversarial patches. This characteristic can be exploited to more accurately identify the location of these patches.}
  \vspace{-0.2cm}
  \label{fig:1}
\end{figure*}

Recently, diffusion models \cite{sohl2015deep,ho2020denoising} have gained prominence as a powerful class of generative models, showing success in enhancing adversarial robustness through the purification of input data \cite{nie2022diffusion,wang2022guided,xiao2022densepure}. Initially, we hypothesize that diffusion purification might be effective in defending against patch attacks. However, our experiments reveal that this approach falls short, as it fails to eliminate adversarial patches. Instead, we discover a phenomenon we term \textbf{Adversarial Anomaly Perception (AAP)}, illustrated in Fig.~\ref{fig:1}. This phenomenon demonstrates that by analyzing the differences between multiple denoised versions of an image, it is possible to localize adversarial patches. This insight allows for the targeted restoration of the specific regions affected by the patch. The underlying reason for this phenomenon may be that adversarial patches are often intricately designed perturbations or contextually inappropriate elements that starkly contrast with the natural image distributions on which the model was trained. This discovery advances our understanding of how diffusion models can differentially respond to adversarial patches, thereby addressing the challenge of balancing the removal of patches with the preservation of image semantics.

Based on the AAP phenomenon, we further introduce \textbf{DIFFender}, a novel \textbf{DIF}fusion-based De\textbf{Fender} framework against adversarial patch attacks, utilizing the text-guided diffusion models. DIFFender operates by first localizing the adversarial patch through the analysis of discrepancies among various denoised images and then restoring the affected region while maintaining the integrity of the original content. These two stages are guided by a unified diffusion model, which allows for a synergistic interaction that enhances overall defense performance—precise localization aids in effective restoration, and successful restoration, in turn, validates and refines the localization process. To further improve accuracy, we integrate a text-guided diffusion model, enabling DIFFender to leverage textual prompts for more precise localization and recovery of adversarial patches. Additionally, we develop a few-shot prompt-tuning algorithm that simplifies and accelerates the tuning process, allowing the pre-trained diffusion model to seamlessly adapt to the adversarial defense task, thereby enhancing robustness. 

Another advantage of DIFFender is its ability to easily extend adversarial patch defense tasks into other domains, such as infrared data. When transferring to the infrared domain, the primary challenges we face include two aspects: the first is the domain shift encountered by the diffusion model when moving from visible to infrared data, and the second is the unique characteristics of infrared data, the weaker texture in infrared images. To meet these two challenges, we design an \textbf{Infrared Domain Constrained (IDC) Token} to improve the prompt, addressing the domain shift problem of infrared data. Additionally, we introduce two more loss functions for prompt tuning: \textbf{Temperature Non-uniformity Correction Loss} and \textbf{Infrared Edge-aware Loss}, to address the issue of unique characteristics in infrared images. This allows DIFFender to seamlessly transition to adversarial patch defense tasks in the infrared domain, making it the first defense method targeting infrared patches. Experimental results show that DIFFender provides strong defense capabilities against infrared adversarial patches under digital and physical settings. The DIFFender pipeline is depicted in Fig.~\ref{fig:2}, and its code is available in \url{ https://github.com/kkkcx/DIFFender}.

In summary, our contributions are as follows:

\begin{itemize}

\item We reveal the Adversarial Anomaly Perception (AAP) phenomenon within diffusion models, which allows for the precise localization of adversarial patches by exploiting the distributional discrepancies between these patches and natural images. This discovery effectively resolves the trade-off between removing adversarial patches and preserving image semantics, expanding the utility of diffusion models.

\item Building on the AAP phenomenon, we introduce DIFFender, an innovative defense framework based on diffusion models. DIFFender utilizes a single diffusion model to both localize and restore adversarial patches, integrating an efficient prompt-tuning module and novel loss functions to jointly train the framework. To our knowledge, DIFFender is the first framework to fully leverage diffusion models for comprehensive defense against patch attacks.

\item We conduct extensive experiments across image classification, face recognition, and real-world scenarios, demonstrating that DIFFender significantly reduces attack success rates, even against strong adaptive attacks. Our results also show that DIFFender generalizes well across various scenarios, different classifiers, and multiple attack methods.

\item Furthermore, we extend DIFFender to the infrared domain, which allows for the multi-modal defense against infrared or visible patch attacks via a unified framework. To the best of our knowledge, DIFFender is the first defense method under this scene, expanding the application scope of patch defense.
\end{itemize}

This work is an extension of our ECCV version~\cite{kang2023diffender}, with major improvements aimed at broadening its applications. Specifically, there are four key enhancements. \textbf{First}, conceptually, we extend the proposed DIFFender method to patch defense in the infrared domain, enabling  multi-modal (infrared or visible) defense. \textbf{Second}, we discuss related works on multi-modal physical defense and attacks, asserting that DIFFender is the first defense method targeting both visible and infrared patch attacks (Sec.~\ref{sec:Related}). \textbf{Third}, methodologically, to address the domain shift issue when transferring to infrared data and the challenge of weaker textures in the infrared domain, we design the IDC token to improve the original prompt and introduce two new loss functions for prompt tuning (Sec.~\ref{sec:Extension_Infrared}). \textbf{Fourth}, experimentally, we evaluate DIFFender’s defense performance against various infrared adversarial patch attacks on both one-stage and two-stage object detectors (Sec.~\ref{sec:Experiments_Infrared}). Additionally, we validate DIFFender's defense capabilities against infrared patches in physical-world experiments to comprehensively test its performance (Sec.~\ref{sec:Experiments_Infrared_physical}), demonstrating its effectiveness for infrared patch defense tasks. 

The rest of this paper is organized as follows: related works are discussed in Sec. \ref{sec:Related}. We introduce our DIFFender in Sec. \ref{sec:Methodology} and Sec. \ref{sec:Extension_Infrared} for visible domain and infrared domain, respectively. Experiments are conducted in Sec. \ref{sec:Experiment} and Sec. \ref{sec:Experiments_Infrared}.  The conclusion is given in Sec. \ref{sec:Conclusion}.

\section{Related Works} \label{sec:Related}

\subsection{Adversarial Attacks} 
Deep neural networks (DNNs) can be deceived into producing incorrect outputs by introducing small perturbations to the input data. Most adversarial attacks \cite{goodfellow2014explaining,moosavi2016deepfool,madry2017towards,dong2018boosting} achieve this by subtly altering pixel values, leading to misclassification  errors. While these techniques are effective in generating adversarial examples in digital environments, they often lack practicality in real-world applications.

In contrast, adversarial patch attacks mislead models by applying a visible pattern or sticker to a localized area of an object, a method that is more feasible in physical settings. First introduced in \cite{brown2017adversarial}, adversarial patches target deep neural networks used in real-world scenarios, aiming to introduce unbounded perturbations within specific regions of clean images. Unlike $\ell_p$-norm-based perturbations, which are designed to be imperceptible, adversarial patches are conspicuous yet resilient modifications, making them particularly effective for physical attacks. These patches have been widely employed across various visual tasks, posing significant threats to model deployment.

Previous research has explored various approaches to developing more effective patches. For example, meaningless patch attacks like LaVAN \cite{karmon2018lavan} randomly select patch locations and generate perturbations, while GDPA \cite{li2021generative} optimizes both patch placement and content to enhance attack effectiveness. Similarly, Wei et al.~\cite{wei2022simultaneously} introduced a reinforcement learning framework to jointly optimize texture and position in a black-box setting. Additionally, RHDE \cite{wei2022adversarial} proposed a natural and practical patch attack method, using real stickers and optimizing their placement for adversarial purposes. This approach not only achieves a high success rate but is also easy to implement, as it can utilize common materials like cartoon stickers as fixed patterns.

\subsection{Adversarial Defenses}
As adversarial attacks have evolved, numerous defense mechanisms have been proposed. However, most existing defenses predominantly address global perturbations constrained by $\ell_p$ norms, including earlier diffusion-based defenses \cite{nie2022diffusion,wang2022guided,xiao2022densepure}, while defenses specifically targeting patch attacks have received less attention. Although adversarial training \cite{wu2019defending,rao2020adversarial} and certified defenses \cite{gowal2019scalable,chiang2020certified} are effective against certain types of attacks, they often fail to generalize to other forms of patch attacks.

As a result, many studies have focused on pre-processing defenses. For example, Digital Watermarking \cite{hayes2018visible} employs saliency maps to detect adversarial regions and uses erosion operations to eliminate small perturbations. Local Gradient Smoothing \cite{naseer2019local} targets regions with high gradient amplitudes, smoothing gradients to mitigate the high-frequency noise introduced by patch attacks. Feature Normalization and Clipping \cite{yu2021defending} reduces informative class evidence by performing gradient clipping, leveraging network structure knowledge. Jedi \cite{tarchoun2023jedi} uses entropy-based masking, while SAC \cite{liu2022segment} offers a general framework for detecting and removing adversarial patches.

While these methods offer some defense against patch attacks, they often struggle to accurately reconstruct the original image and can be circumvented by adaptive attacks \cite{athalye2018obfuscated}. In contrast, we propose utilizing pre-trained diffusion models to more precisely localize and restore adversarial patches. Our approach enables the accurate identification of adversarial patch locations during the localization stage by exploiting the inherent properties of diffusion models. During the restoration stage, the diffusion model reconstructs the affected regions while preserving the visual integrity of the image. Notably, these two stages are guided by a unified diffusion model, allowing for synergistic interaction that enhances overall defense effectiveness. Additionally, we introduce a few-shot prompt-tuning algorithm to fine-tune the diffusion model, ensuring that the pre-trained model seamlessly adapts to the defense task.

\subsection{Infrared Adversarial Attacks and Defenses}
Adversarial examples are prevalent across various domains. Recently, researchers have begun to explore adversarial examples in infrared imagery. Edwards and Rawat~\cite{edwards2020study} investigated the performance of adversarial attacks in ship detection under thermal infrared imaging. Osahor and Nasrabadi~\cite{osahor2019deep} explored how to generate visually imperceptible adversarial infrared examples that can evade detection by deep neural network-based object detectors. These methods generate perturbations by altering pixel values within infrared images, thus rendering them impractical for use in the physical world. To address this issue, Zhu et al.~\cite{zhu2021fooling} made the first attempt to create physical adversarial examples using a set of small bulbs that modify the infrared radiation distribution of an object by simulating additional heat sources. Subsequently, Zhu et al.~\cite{zhu2022infrared} proposed adversarial clothing designed to deceive infrared detectors from various angles by enveloping the entire body. Wei et al.~\cite{wei2023unified} introduced a method named Unified Adversarial Patch (UAP), which designs a unified adversarial patch capable of affecting detection systems across different modalities. Specifically, the authors constructed a patch that produces adversarial effects in both visible and infrared images, facilitated by the use of special materials and coatings for multi-modal attacks. Furthermore, Wei et al.~\cite{wei2023infrared} proposed Adversarial Infrared Patches, focusing on designing the shape and location of patches rather than complex patterns, making them easy to implement in physical world.

In the defense domain, techniques such as PixelMask~\cite{agarwal2021cognitive}, Bit squeezing~\cite{xu2017feature}, JPEG compression~\cite{dziugaite2016study}, Spatial Smoothing~\cite{xu2017feature}, and Total variation minimization~\cite{guo2017countering} are employed to defend against infrared patch attacks. However, these methods were not specifically designed for infrared patch attacks and thus do not achieve satisfactory results. Current research on patch adversarial defense predominantly focuses on the RGB modality, with little attention to others such as the infrared modality. Our work showcases the first to concurrently address both RGB and infrared modalities, further validating multi-modal attack defense.

\begin{figure*}[tp]
  \centering
  \includegraphics[width=0.92\linewidth]{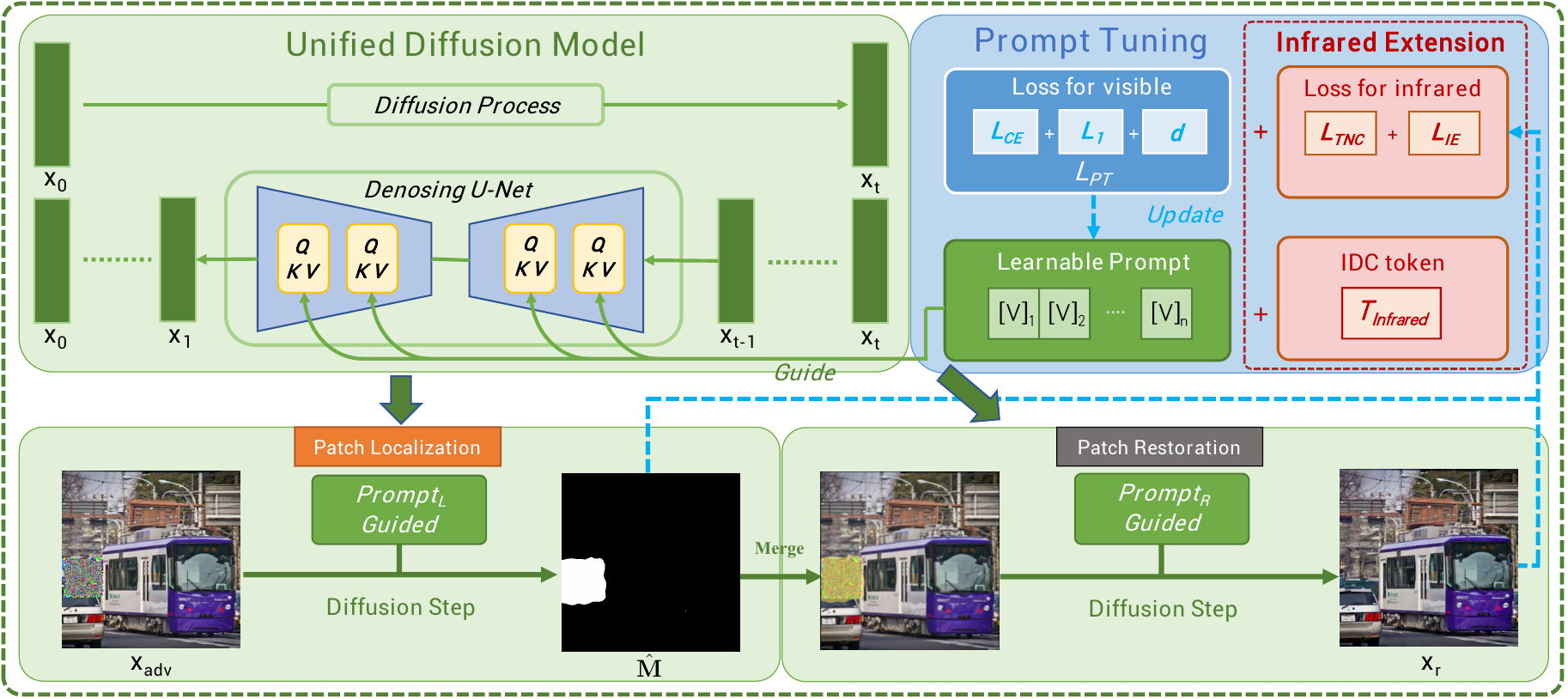}
  \caption{\small Pipeline of DIFFender. 
  DIFFender utilizes a unified diffusion model to seamlessly coordinate the localization and restoration of adversarial patch attacks, integrating a prompt-tuning module to enable efficient and precise tuning.
  }
  \label{fig:2}
\end{figure*}

\begin{figure*}[!h]
  \centering
  \includegraphics[width=0.90\linewidth]{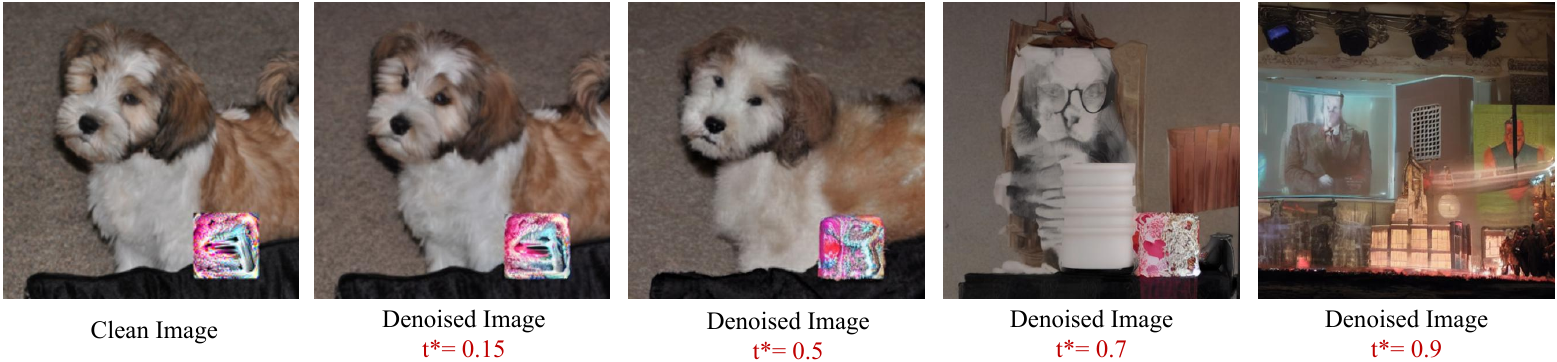}  
 \caption{\small Denoised results at different noise ratios. 
 With smaller ratios (\( t^* = 0.15/0.5 \)), the patch remains unpurified; however, with larger ratios (\( t^* = 0.7/0.9 \)), the global structure is compromised.
 }
  \label{fig:new}
\end{figure*}

\section{Methodology}
\label{sec:Methodology}

In this section, we present the proposed \textbf{DIFFender} framework. The pipeline of DIFFender, illustrated in Fig.~\ref{fig:2}, comprises three key modules: patch localization, patch restoration, and prompt tuning. We begin by discussing the discovery of the Adversarial Anomaly Perception (AAP) phenomenon within diffusion models in Sec.~\ref{sec:4-1}. Building on this insight, we then describe the overall architecture of DIFFender in Sec.~\ref{sec:4-2}, followed by an in-depth explanation of the enhanced techniques introduced through prompt tuning in Sec.~\ref{sec:4-3}.

\subsection{Discovery of the AAP Phenomenon}\label{sec:4-1}

\begin{figure*}[h]
  \centering
  \includegraphics[width=0.99\linewidth]{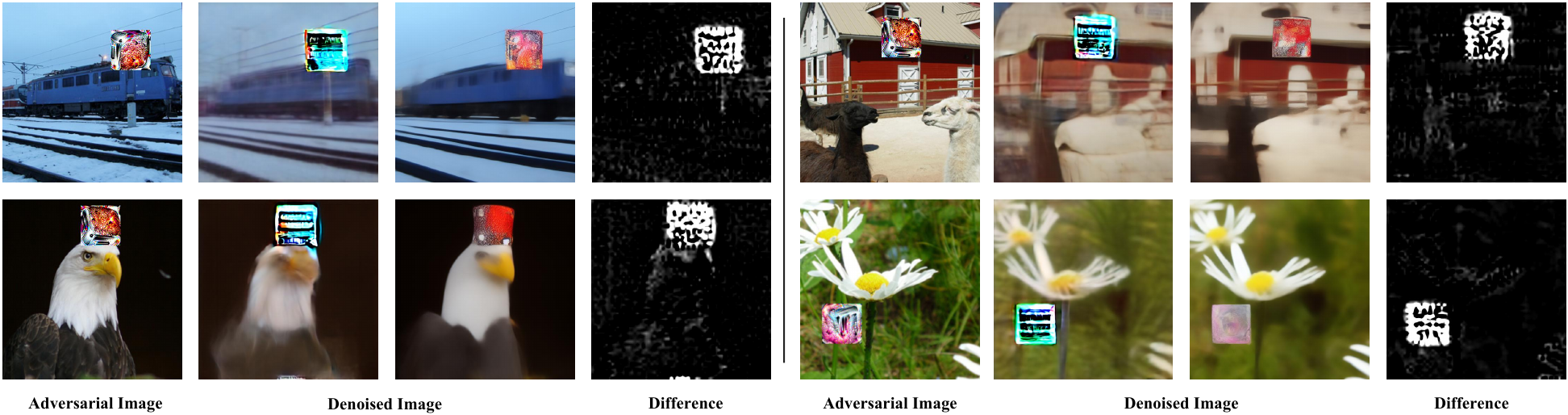}  
\caption{\small In our analysis of ImageNet images, we observe a pronounced difference specifically within regions impacted by adversarial patches, offering empirical evidence in support of the AAP phenomenon.}
  \label{fig:observation}
\end{figure*}

DiffPure \cite{nie2022diffusion} is a recent method that employs diffusion models to remove imperceptible perturbations by introducing Gaussian noise at a predetermined ratio \( t^* \) (ranging from 0 to 1) to adversarial images, followed by a denoising process using the reverse dynamics of diffusion models. Our initial objective was to evaluate the effectiveness of DiffPure against patch attacks. However, as shown in Fig. \ref{fig:new}, our empirical results reveal that DiffPure is insufficient for countering patch attacks. This shortcoming arises from a fundamental trade-off: a larger \( t^* \) is required to effectively purify adversarial perturbations, but this also risks compromising the image’s semantic integrity, whereas a smaller \( t^* \) preserves semantics but fails to eliminate the adversarial patches. This makes it impossible to identify an optimal noise ratio that can effectively defend against patch attacks.

In contrast, we observed that at a critical noise ratio \( t^* \), a unique pattern emerged: while adversarial patches resisted denoising and struggled to be restored, the rest of the image remained semantically intact. This observation suggests that by analyzing differences between various denoised images, it is possible to identify the regions containing adversarial patches. This finding, illustrated in Fig. \ref{fig:observation}, led to the identification of the \textbf{Adversarial Anomaly Perception (AAP)} phenomenon. 

The AAP phenomenon likely occurs because adversarial patches are often carefully engineered perturbations with complexity that far exceeds the natural noise found in real image datasets. Alternatively, they may represent meaningful stickers placed in contextually inappropriate locations, making them stand out as anomalies. Since diffusion models are trained to learn the probability distribution of natural images, they struggle to adapt to the distribution of adversarial examples in their latent space, leading to noticeable discrepancies.

The discovery of AAP offers valuable insights into how diffusion models can differentially respond to adversarial patches. It enables the diffusion model to detect and localize adversarial patches by analyzing distributional discrepancies, which in turn facilitates targeted restoration of the affected areas. This approach effectively resolves the trade-off between eliminating adversarial patches and preserving the authenticity of the image. Building on the AAP phenomenon, we propose \textbf{DIFFender}, a unified diffusion-based defense framework that employs a single diffusion model to both localize and restore patch attacks.

\subsection{DIFFender}\label{sec:4-2}

\noindent\textbf{Patch Localization.}  
DIFFender begins with precise patch localization, leveraging the AAP phenomenon observed in diffusion models. For an adversarial image $\mathbf{x}_{adv}$, we first introduce Gaussian noise to generate a noisy image $\mathbf{x}_t$ with a specific noise ratio \( t^* \) (set to 0.5 in our experiments). Next, a text-guided diffusion model is applied to denoise $\mathbf{x}_t$, producing $\mathbf{x}_p$ using a textual prompt $prompt_{L}$, and $\mathbf{x}_e$ using an empty prompt. The adversarial patch region $\hat{\mathbf{M}}$ is then estimated by calculating the difference between the denoised images $\mathbf{x}_p$ and $\mathbf{x}_e$.

However, diffusion models typically require a significant number of time steps $T$, leading to high computational costs. To mitigate this, we directly predict the image $\mathbf{x}_0$ from $\mathbf{x}_t$ in a single step, reducing the processing time by a factor of $T$. Although the one-step prediction may introduce some blurriness and discrepancies, the differences between one-step predictions still reflect the AAP phenomenon. In practice, we perform one-step denoising twice, yielding two results: $\mathbf{x}_a$, guided by $prompt_{L}$, and $\mathbf{x}_b$, guided by an empty prompt. The difference is then binarized to estimate the patch region as follows:
\begin{equation}
\hat{\mathbf{M}} = \mathrm{Binarize}\left(\frac{1}{m} \sum\limits_{i=0}^m (\mathbf{x}_a^i - \mathbf{x}_b^i)\right),
\end{equation}
where the difference is computed $m$ times (set to 3 in our experiments) to enhance stability and reduce randomness. The $prompt_{L}$ can be manually designed (e.g., "adversarial") or automatically tuned, as discussed in Sec.~\ref{sec:4-3}.

Specifically, we compute the difference between the latent denoising results for each pair of noisy inputs. The absolute differences of the latent variables are summed across channels, averaged, and normalized. 

\begin{figure}[!h]
  \centering
  \includegraphics[width=0.99\linewidth]{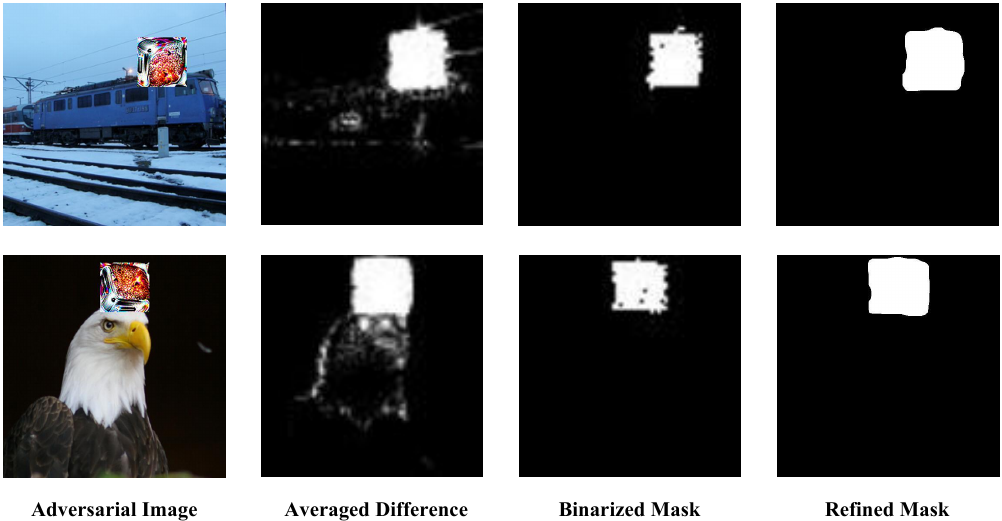} 
  \vspace{-1ex}\caption{\small To refine the mask, the estimated differences are binarized, followed by Gaussian smoothing and dilation operations.}
  \label{fig:refine}
\end{figure}

\noindent\textbf{Mask Refinement.}  
As shown in Fig. \ref{fig:refine}, the initial mask derived from the averaged difference may contain minor inaccuracies. To address this, we first binarize the difference using threshold \( \theta \) to obtain an initial mask, then refine it by sequentially applying Gaussian smoothing and dilation operations. This process yields a more accurate estimation $\hat{\mathbf{M}}$ of the patch region. The resulting mask edges may slightly extend beyond the patch area, ensuring consistency during patch restoration and thereby enhancing the overall performance of the defense pipeline.

\noindent\textbf{Patch Restoration.}
Once the patch region has been localized, DIFFender proceeds to restore the affected area, aiming to eliminate adversarial effects while maintaining the overall coherence and quality of the image. Specifically, we combine the estimated mask $\hat{\mathbf{M}}$ with the adversarial image $\mathbf{x}_{adv}$ as inputs to a text-guided diffusion model, using the prompt $prompt_{R}$ to generate a restored image $\mathbf{x}_r$. This process follows the inpainting pipeline of Stable Diffusion \cite{Rombach_2022_CVPR}, where a UNet model is employed with an additional five input channels to incorporate the estimated mask $\hat{\mathbf{M}}$. As with localization, $prompt_{R}$ can be manually set (e.g., "clean") or automatically tuned.

To ensure the complete removal of adversarial effects, the restoration process involves the full diffusion sequence, which requires more processing time than the localization step. Although this increases the time complexity, DIFFender can be set to mitigate this by invoking the restoration algorithm only when a patch attack is detected. In real-world applications, adversarial attacks typically affect only a small fraction of images. Thus, the more time-intensive restoration (which takes T time steps) is only necessary for a few cases, while the majority of images can be handled by the lightweight localization module (requiring just 1 time step), which benefits from acceleration techniques discussed earlier. As a result, DIFFender significantly reduces the time complexity associated with the diffusion framework, making it several times more efficient than DiffPure, which is also diffusion-based. This efficiency improvement makes DIFFender a practical solution for real-world deployment.

\noindent\textbf{Unified Defense Model.}
The two stages described above—patch localization and patch restoration—are seamlessly integrated into a unified diffusion model (e.g., Stable Diffusion), guided by the critical AAP phenomenon. This deliberate integration allows us to capitalize on the close interaction between these stages, significantly enhancing the overall defense mechanism. Building on these insights, we have also introduced a prompt-tuning module that optimizes the entire pipeline as a cohesive unit.

\subsection{Prompt Tuning}\label{sec:4-3}

In line with the pipeline described, DIFFender leverages pre-trained diffusion models to enable efficient zero-shot defense. While this approach is effective in most instances, slight inaccuracies in the segmented masks can occasionally occur in certain challenging cases. Given that vision-language pre-training harnesses the power of large-capacity text encoders to explore an extensive semantic space~\cite{zhou2022learning}, we introduce the prompt-tuning algorithm to effectively adapt these learned representations for adversarial defense tasks by only few-shot tuning.

\noindent\textbf{Learnable Prompts.}
We begin by replacing the textual vocabulary with learnable continuous vectors. 
Unlike text prompts, learnable prompts are a set of continuous vectors that allow for gradient computation to identify the most effective prompt. 
In this way, $prompt_{L}$ and $prompt_{R}$ are represented as vectors as follows:
\begin{equation}
\begin{split}
prompt_{L} = [V_L]_1 [V_L]_2 \ldots [V_L]_n; \\
prompt_{R} = [V_R]_1 [V_R]_2 \ldots [V_R]_n,
\end{split}
\end{equation}
where each $[V_L]_i$ or $[V_R]_i$ ($i \in \{1, \ldots, n\}$) is a vector of the same dimensionality as word embeddings. The hyperparameter $n$ defines the number of context tokens, which we set to 16 by default. The initial content of $prompt_{L}$ and $prompt_{R}$ can be manually specified or randomly initialized.

\noindent\textbf{Tuning Process.}
Once the learnable vectors are established, we introduce three loss functions to guide prompt tuning. These losses are designed to jointly optimize the vectors, enhancing the model's ability to identify adversarial regions and improving overall defense performance.

First, to accurately localize adversarial regions, we employ a cross-entropy loss that compares the estimated mask $\hat{\mathbf{M}}$ with the ground-truth mask $\mathbf{M}$:
\begin{equation}
 L_{CE}(\mathbf{M},\hat{\mathbf{M}}) = -\sum_{i=1}^d\mathbf{M}_i\log(\hat{\mathbf{M}}_i),
\end{equation}
where $i$ refers to the $i$-th element of the mask.
Next, in the patch restoration module, our goal is to restore the affected region while eliminating the adversarial impact. We achieve this by calculating the $\ell_1$ distance between the restored image $\mathbf{x}_r$ and the clean image $\mathbf{x}$:
\begin{equation}  
L_1(\mathbf{x}_r, \mathbf{x})=\left| \mathbf{x}_r-\mathbf{x} \right|.
\end{equation}
Finally, to ensure the adversarial effects are fully mitigated, we draw inspiration from \cite{liao2018defense} and \cite{zhang2018unreasonable} by aligning the high-level feature representations of the restored image $\mathbf{x}_r$ and the clean image $\mathbf{x}$. Specifically, we compute the $\ell_2$ distance between their feature representations, weighted by a layer-wise hyperparameter:
\begin{equation}
d\left(\mathbf{x}_r, \mathbf{x}\right)=\sum_l \frac{1}{H_l W_l} \sum_{h, w}\left\|w_l \odot\left(\hat{y}_{r h w}^l-\hat{y}_{c h w}^l\right)\right\|_2^2,
\end{equation}
where $l$ denotes a specific layer in the network, $\hat{y}_r^l,\hat{y}_c^l \in \mathcal{R}^{H_l\times W_l \times C_l}$ are the unit-normalized results across the channel dimension, and the vector $w^l\in\mathcal{R}^{C_l}$ is used to scale activation channels.

The overall loss function $L_{PT}$ for prompt tuning is then obtained by summing the three losses:
\begin{equation}
L_{PT} = L_{CE}(\mathbf{M},\hat{\mathbf{M}}) + L_1(\mathbf{x}_r, \mathbf{x}) + d\left(\mathbf{x}_r, \mathbf{x}\right).
\end{equation}
We minimize $L_{PT}$ w.r.t. $prompt_L$ and $prompt_R$ using gradient descent. The continuous representation design facilitates thorough exploration of the embedding space.

\noindent\textbf{Few-Shot Learning.}
During prompt tuning, DIFFender leverages a limited set of images for few-shot learning. Specifically, the model is fine-tuned on a limited number of attacked images (8-shot in our experiments) from a single attack type, enabling it to learn optimal prompts that generalize effectively across different scenarios and attacks. This approach ensures the tuning module is both efficient and straightforward.

\section{Extension to the Infrared Domain}\label{sec:Extension_Infrared}

In this section, we explore how to adapt our DIFFender to defend against infrared adversarial patches. 

As previously mentioned, DIFFender is capable of localizing adversarial patches and subsequently employing a restoration module to repair adversarial patches. Given that infrared domain patch attacks follow a similar paradigm to RGB patch attacks, DIFFender possesses the potential to extend to infrared patch defense. However, transitioning to the infrared domain presents two primary challenges: (1) the domain transfer issue when the diffusion model moves from visible to infrared data, where the diffusion model is trained using dataset like LAION-5B, ensuring generalization across different scenes in the visible spectrum, with infrared images only constituting a small portion; (2) the inherent differences between infrared and RGB images, which lead to notable discrepancies in the patches generated for infrared attacks, such as the richer textures and colors in RGB images compared to the weaker textures in infrared images. These two issues make the method in Sec. \ref{sec:Methodology} not directly deal with infrared adversarial patches.

Building on Sec. \ref{sec:Methodology}, Sec. \ref{sec:Extension_Infrared_1} introduces the Infrared Domain Constrained Token (IDC token) to enhance the prompt. Sec. \ref{sec:Extension_Infrared_2} presents two new loss functions for prompt tuning, and Sec. \ref{sec:Extension_Infrared_3} details the prompt tuning process for the infrared domain.

\subsection{Infrared Domain Constrained Token} \label{sec:Extension_Infrared_1}
  \vspace{0.2cm}
\begin{figure}[!h]
  \centering
  \includegraphics[width=0.99\linewidth]{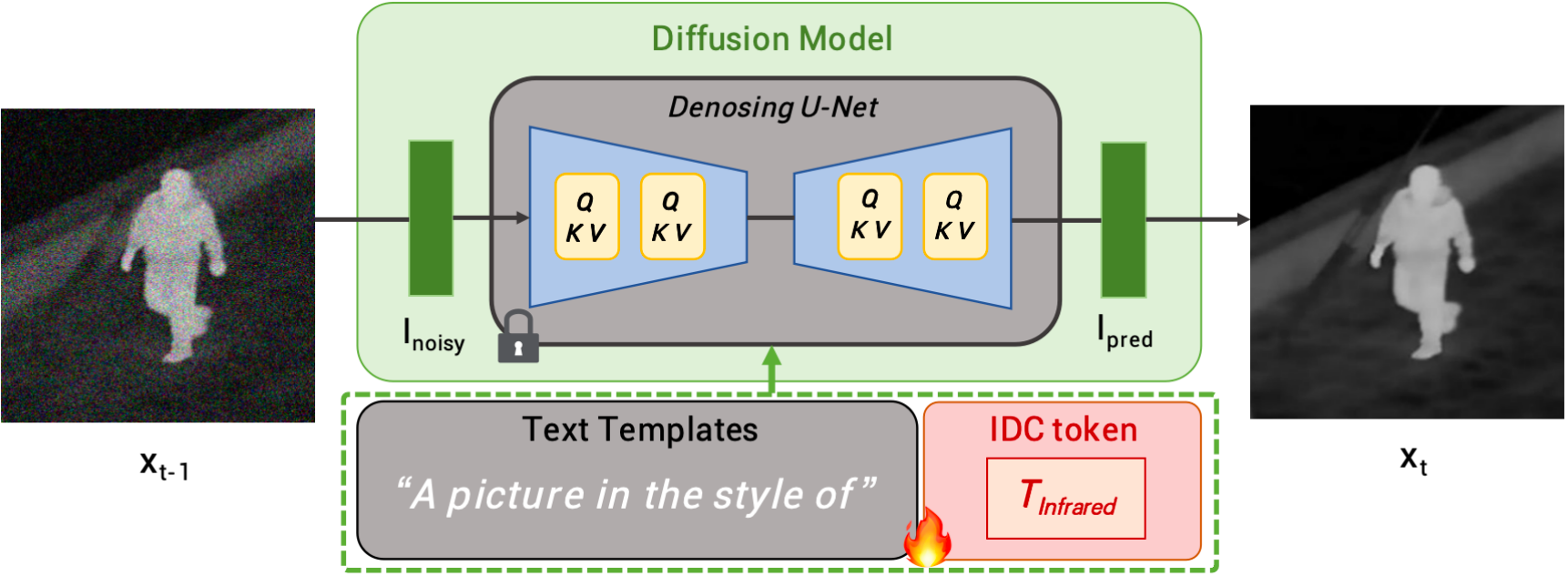} 
  \vspace{-1ex}\caption{\small The training process of IDC token. The weights of the pre-trained diffusion model are frozen in the training process.}
  \label{fig:IDC_training}
\end{figure}
  \vspace{-0.2cm}

Diffusion models are trained using large-scale datasets to ensure generalization across various scenes in the visible spectrum, though infrared images constitute only a minor portion. To better adapt the diffusion model for infrared defense tasks and address the domain transfer issues from visible to infrared data, we designed the "Infrared Domain Constrained Token" (IDC token).

\begin{figure}[!t]
  \centering
  \includegraphics[width=0.90\linewidth]{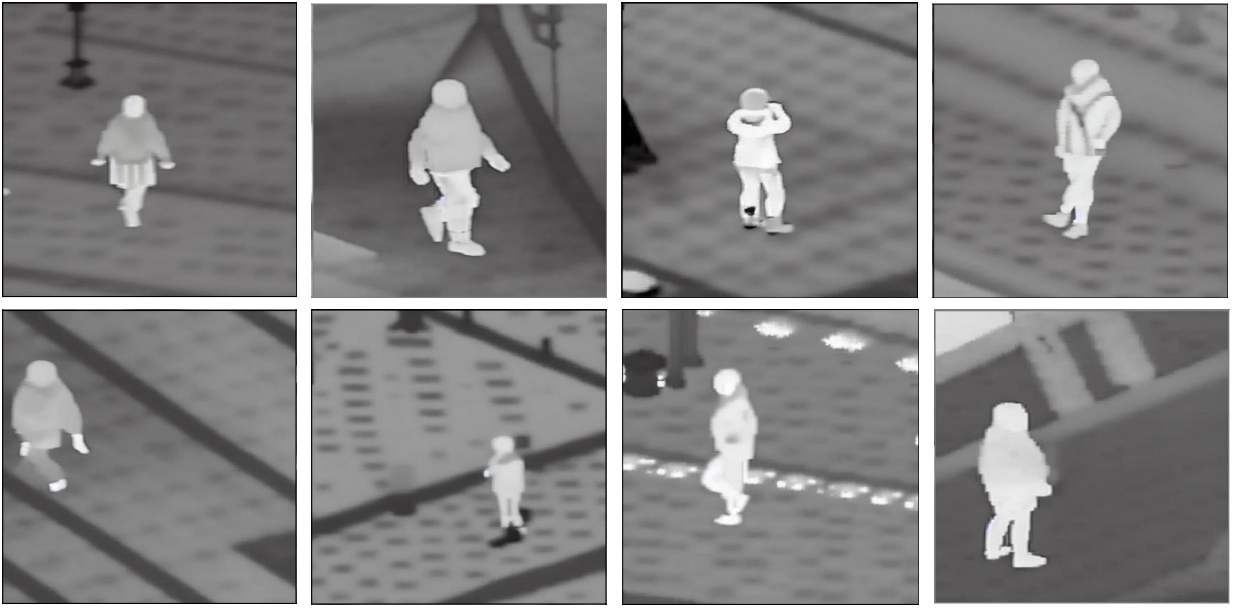} 
  \vspace{-1ex}\caption{\small The generated images constrained by IDC token. IDC token ensures that the diffusion model only produces images in the infrared domain.}
  \label{fig:infrared_gen}
  \vspace{-2ex}
\end{figure}

Given a small set of specific images captured by an infrared camera (10 images used in the experiments), the goal of the IDC token is to identify a token \( T_{\text{Infrared}} \) in the diffusion model's textual space that accurately captures the concept of the infrared domain. We use a series of fixed text templates in the training phase, such as "a rendering in the style of \( T_{\text{Infrared}} \)", "a picture in the style of \( T_{\text{Infrared}} \)",  to guide the diffusion model in capturing the concept. The result \( T_{\text{Infrared}} \) will help the diffusion model constrain the generated output within the infrared domain, as shown in Fig. \ref{fig:infrared_gen}. This token is then concatenated to the DIFFender prompt and remains frozen during prompt tuning, as Fig. \ref{fig:2}. 
This design extends the Adversarial Anomaly Perception (AAP) to infrared patch attack localization and ensures the restoration module correctly restores adversarial patch areas, it also guarantees that removing \( T_{\text{Infrared}} \) doesn't impact DIFFender's performance in the visible spectrum.

Specifically, to create such a customized token \( T_{\text{Infrared}} \), we learn the corresponding embedding vector \( E_{\text{Infrared}} \) (768 dimensions in the experiment) within the text embedding space of the diffusion model. To learn \( E_{\text{Infrared}} \), we freeze the weights of the Encoder and UNet in the entire pre-trained diffusion model and find the \( E_{\text{Infrared}} \) that minimizes the original training objective of the diffusion model, as follows:
\begin{equation}
   E_{\text{Infrared}} = \argmin_v \mathbb{E}_{z\sim\mathcal{E}(x), y, \epsilon \sim \mathcal{N}(0, 1), t }\Big[ \Vert \epsilon - \epsilon_\theta(z_{t},t, c_\theta(y)) \Vert_{2}^{2}\Big] \, ,
\end{equation}
This is achieved by reusing the same training scheme as the diffusion model, where \( c_\theta(y) \) maps the conditioning input \( y \) to a conditioning vector, \( t \) is the time step, \( z_t \) represents the latent noise at time \( t \), \( \epsilon \) is the noise sample, and \( \epsilon_\theta \) is the denoising network, while keeping \( c_\theta \) and \( \epsilon_\theta \) fixed, as shown in Fig. \ref{fig:IDC_training}. 

\subsection{Loss Functions for Infrared Domain} \label{sec:Extension_Infrared_2}

\noindent\textbf{Temperature Non-uniformity Correction Loss.}
Observations indicate that existing infrared adversarial attack methods utilize the application of thermal insulating materials to create non-uniform infrared signals, thereby disrupting the perception of infrared sensors, typically manifesting as patches or spots in images. Based on this observation, infrared adversarial patch attacks result in non-uniformity in infrared images. Consequently, we have designed a Temperature Non-uniformity Correction (TNC) Loss to mitigate the effects of non-uniformity. This loss function not only measures the temperature non-uniformity in the infrared images to represent the adversarial content but also evaluates the differences between the restored image and the clean image, considering the accuracy and consistency of temperature information.

Assuming \( I \) is the clean infrared image and \( I_r \) is the corrected image, the TNC Loss can be designed to include two main components: one for measuring the uniformity of the corrected image, and another for preserving image details. The model can be expressed as follows:
\begin{equation}
L_{TNC} = \alpha \cdot L_{uniform}(I_{r}) + \beta \cdot L_{detail}(I, I_{r}),
\end{equation}
where \( \alpha \) and \( \beta \) are hyperparameters used to balance these two aspects. The uniformity loss \( L_{\text{uniform}} \) uses the local standard deviation to quantify the uniformity of temperature in the corrected image. A lower local standard deviation indicates a more uniform image area, thus a reduced adversarial effect:
\begin{equation}
L_{\text{uniform}} = \frac{1}{N} \sum_{i,j} \sqrt{\sum_{k \times k} (p_{ij} - \mu_{ij})^2},
\end{equation}
where \( k \times k \) is the pixel neighborhood window, with \( k \) being an odd number to ensure a central pixel in the window. For each pixel \( p_{ij} \) in image \( I_r \), the average value \( \mu_{ij} \) of pixel values within its \( k \times k \) neighborhood is calculated. The sum of squared differences between the pixel values and the neighborhood average \( \mu_{ij} \) is computed, then the square root is taken to obtain the local standard deviation. Finally, the mean of all pixel's local standard deviations is calculated to yield the overall image uniformity loss \( L_{\text{uniform}} \).

The detail preservation loss \( L_{\text{detail}} \) aims to ensure that the correction process does not compromise important image details. This can be achieved by using a similarity measure between the clean infrared image and the restored infrared image, where we employ the Structural Similarity Index (SSIM):
\begin{equation}
L_{\text{detail}} = \text{SSIM}(I, I_r).
\end{equation}

\noindent\textbf{Infrared Edge-aware Loss.}
Compared to RGB images, infrared images often lack textures and colors, emphasizing edge information instead. To enhance the model's ability to recognize edges in infrared imagery, we have designed the Infrared Edge-aware Loss, which underscores the importance of edge regions in the perception of infrared images. The edge-aware loss is implemented by first using an edge detector \( E \) to extract edge information from the clean image \( I \), and then calculating the differences between the model prediction \( I_r \) and \( I \) in edge regions, as well as in non-edge regions, incorporating a balancing weight that assigns greater importance to prediction errors in edge areas.

Assuming \( I_r \) is the image restored by the defense method and \( I \) is the clean image, the Infrared Edge-aware Loss \( L_{IE} \) can be expressed as:
\begin{equation}
L_{IE} = \gamma \cdot L_{D}(E(I), E(I_r)) + \delta \cdot L_{D}((1-E(I)), (1-E(I_r)))
\end{equation}
where \( L_D \) is the Dice loss used to quantify the differences between the processed restored image and the clean image. \( \gamma \) and \( \delta \) are hyperparameters used to balance the losses in edge and non-edge areas, respectively. \( E() \) denotes the applied edge detector, which outputs a binary map of the edge regions (Sobel operator in experiments).

The Infrared Edge-aware Loss emphasizes the importance of edge recognition, enabling  DIFFender to adapt to infrared domain patch attacks and helping the model achieve better performance in infrared image defense tasks.

\subsection{Prompt Tuning for Infrared Domain}\label{sec:Extension_Infrared_3}
As shown in Fig. \ref{fig:2}, prompt tuning in the infrared domain mirrors the approach used for the visible domain, with the key difference being the concatenation of a pre-trained IDC token, which remains frozen during tuning. For the loss function, in addition to the original visible domain losses, \( L_{TNC} \) and \( L_{IE} \) are added to better adapt to the infrared domain. DIFFender is fine-tuned on a limited set of attacked images (8-shot) from a single attack. The hyperparameters \( \alpha \), \( \beta \), \( \gamma \), and \( \delta \) are set to 0.4, 0.6, 0.7, and 0.3, respectively.

\section{Experiments in the Visible Domain}
\label{sec:Experiment}
In this section, we give the experiments about ablation study, evaluations on image classification, face recognition, and physical world, etc. Due to space limitations, more experiments can be found in the Appendix. 

\vspace{-0.2cm}

\subsection{Experimental Settings}\label{sec:5-1}

\noindent\textbf{Datasets and Baselines.}
We evaluate our approach on the ImageNet dataset \cite{deng2009imagenet}, comparing it against eight state-of-the-art defense methods. These include image smoothing-based defenses such as JPEG \cite{dziugaite2016study} and Spatial Smoothing \cite{xu2017feature}; image completion-based defenses like DW \cite{hayes2018visible}, LGS \cite{naseer2019local}, and SAC \cite{liu2022segment}; feature-level suppression defense FNC \cite{yu2021defending}; and Jedi \cite{tarchoun2023jedi}, which relies on entropy-based defense. Additionally, we assess the diffusion purification method, DiffPure \cite{nie2022diffusion}. For classification tasks, we utilize two advanced classifiers trained on ImageNet: the CNN-based Inception-v3 \cite{szegedy2016rethinking} and the Transformer-based Swin-S \cite{liu2021swin}.

\noindent\textbf{Adversarial Attacks.}
We evaluate our defense against several adversarial attacks, including AdvP \cite{brown2017adversarial} and LaVAN \cite{karmon2018lavan}, which randomly select patch positions and optimize them; GDPA \cite{li2021generative}, which optimizes both patch position and content; and RHDE \cite{wei2022adversarial}, a natural-looking attack that uses realistic stickers and searches for their optimal placement. To implement adaptive attacks, we use BPDA \cite{athalye2018obfuscated} to approximate gradients, leading to stronger attacks such as BPDA+AdvP and BPDA+LaVAN, effectively making the defense methods white-box against these attacks. Each attack iteration is set to 100, with a patch size of 5\% of the input image. When adapting the attack for DIFFender, we apply an additional Straight-Through Estimator (STE) \cite{yin2019understanding} during backpropagation through thresholding operations.

\noindent\textbf{Evaluation Metrics.}
We assess defense performance using both standard accuracy and robust accuracy metrics. Given the computational demands of adaptive attacks, we evaluate robust accuracy on a subset of 512 images sampled from the test set, unless otherwise noted. To ensure meaningful comparisons, the selected subset consists of images that are correctly classified.

\subsection{Ablation Studies}\label{sec:5-2}
\vspace{-0.1cm}

\begin{table}[h]
 \centering
\renewcommand\arraystretch{1.2}
\caption{\small Ablation study for different loss functions of DIFFender.}
  \scalebox{0.96}{
        \begin{tabular}{ccc|ccccc} 
            \toprule
                   \multicolumn{1}{c}{} & \multicolumn{1}{c}{} & \multicolumn{1}{c}{} & \multicolumn{5}{c}{Inception-v3}        \\ \hline
                $L_{CE}$ & $L_1$& $d$& Clean & AdvP & LaVAN & GDPA & RHDE \\ \hline
                 & \checkmark & \checkmark & \textbf{91.8} & 76.2 & 66.0 & 72.3 &49.2 
\\ 
                \checkmark & ~ & \checkmark & 88.3 & 87.1 & 69.5 & 73.8 &52.7 
\\ 
                \checkmark & \checkmark & ~ & 90.2 & 87.1 & 69.1 & 73.0 &52.0 
\\ 
                \checkmark & \checkmark & \checkmark  & 91.4 & \textbf{88.3} & \textbf{71.9} &  \textbf{75.0} &\textbf{53.5} \\ 
            \bottomrule
          \end{tabular}
         }
          \label{tab:loss}
          \vspace{-0.3cm}
\end{table}

\begin{table}[h]
\centering
\caption{\small Accuracy against attacks of varying patch sizes by Inception-v3.}
\scalebox{1.1}{
\begin{tabular}{c|ccccc}
\toprule
Size & 0.5\% & 1.0\%   & 5.0\%   & 10.0\%  & 15.0\%  \\ \midrule
Undefended & 64.3 & 50.8   & 0.0   & 0.0  & 0.0    \\
SAC \cite{liu2022segment} & 81.8 & 83.8   & 84.2   & 60.9  & 34.8   \\
Jedi \cite{tarchoun2023jedi} & 61.7 & 56.4   & 67.6   & 42.2  & 33.8   \\
DIFFender & \textbf{86.1} & \textbf{87.3}   & \textbf{88.3}   & \textbf{70.5}  & \textbf{56.6}   \\ \bottomrule
\end{tabular}
}
  \label{Patch}
\end{table}

\begin{table*}[!t]
\small
 \centering
\caption{\small Ablation study for restoration modules in DIFFender. "NR" denotes "No Restoration Process". }

  \scalebox{0.99}{
  \begin{tabular}{c|ccccc|ccccc}
  \toprule
               & \multicolumn{5}{c}{Inception-v3}   & \multicolumn{5}{c}{Swin-S}         \\
\textbf{Defense}        & Clean & AdvP & LaVAN & GDPA & RHDE & Clean & AdvP & LaVAN & GDPA & RHDE \\ \midrule
DIFFender (NR) & 86.3 & 84.0 & 66.8 & 69.5 & 48.0 & 88.7 & 92.2 & 81.8 & 78.9 & 69.1 \\
DIFFender & \textbf{91.4} & \textbf{88.3} &\textbf{71.9} & \textbf{75.0} & \textbf{53.5}& \textbf{93.8}& \textbf{94.5} & \textbf{85.9} & \textbf{82.4} & \textbf{70.3}  \\

\bottomrule
\end{tabular}
}
  \label{tab:Restoration}
  \vspace{-0.2cm}
\end{table*}

\begin{table*}[!t]
 \centering
\small
\caption{\small Ablation study for different prompt forms. "EP" and "MP" represent "Empty Prompt" and "Manual Prompt".}
  \scalebox{0.99}{
  \begin{tabular}{c|ccccc|ccccc}
  \toprule
               & \multicolumn{5}{c}{Inception-v3}   & \multicolumn{5}{c}{Swin-S}         \\
\textbf{Defense}        & Clean & AdvP & LaVAN & GDPA & RHDE & Clean & AdvP & LaVAN & GDPA & RHDE \\ \midrule
DIFFender (EP)& 89.1  & 76.4 & 66.8  & 71.1 & 47.0 & 93.2  & 89.8 & 81.4  & 79.3 & 65.7 \\
DIFFender (MP) & 87.3  & 77.9 & 68.2  & 70.3 & 47.8& 92.2  & 91.2 & 82.4  & 77.0 & 67.6 \\
DIFFender & \textbf{91.4} & \textbf{88.3} &\textbf{71.9} & \textbf{75.0} & \textbf{53.5} & \textbf{93.8} & \textbf{94.5} & \textbf{85.9} & \textbf{82.4} & \textbf{70.3} \\
\bottomrule
\end{tabular}
}
  \label{tab:6}
  \vspace{-0.2cm}
\end{table*}

\noindent\textbf{Impact of Loss Functions.}
To assess the impact of different loss functions, we conduct tuning experiments where we remove each loss function—$L_{CE}$, $L_1$, and $d$—individually. The results, presented in Tab.~\ref{tab:loss}, reveal that excluding $L_{CE}$ leads to a significant decrease in robust accuracy, despite an improvement in clean accuracy. This drop occurs because optimizing the restoration module alone, without considering $L_{CE}$, impairs localization performance. Conversely, removing $L_1$ results in a noticeable decline in clean accuracy, as the images are not adequately restored. Omitting either $d$ or $L_1$ also causes a slight reduction in robust accuracy. Overall, DIFFender, which includes all three loss functions, achieves the highest robust accuracy, highlighting the importance of joint optimization and the close interaction between the two modules for maximizing performance.

\noindent\textbf{Impact of Patch Size.}
We conduct experiments to evaluate DIFFender’s performance against adversarial patches of various sizes, using patches generated by AdvP ranging from 0.5\% to 15\% of the image size. These results are compared with those of SAC and Jedi, the state-of-the-art methods. As shown in Tab.~\ref{Patch}, DIFFender demonstrates superior generalization across different patch sizes, benefiting from vision-language pre-training. In contrast, Jedi and SAC are more sensitive to changes in patch size. Notably, DIFFender was only prompt-tuned for patches of 5.0\% size.

\noindent\textbf{Impact of the Restoration Module.}
To determine the necessity of the restoration, we conduct an experiment where the patch restoration step was removed, and the value in the $\hat{\mathbf{M}}$ region was set to zero. The results, displayed in Tab.~\ref{tab:Restoration}, indicate that the inclusion of the restoration step significantly enhances DIFFender’s performance. This improvement is because patches can sometimes obscure critical areas of an image, leading to a loss of semantic information. The restoration step recovers these lost semantics, enabling classifiers to better handle challenging scenarios. Additionally, longer diffusion steps introduce more randomness, which helps maintain accuracy against adaptive attacks. Thus, the restoration module is essential for optimal performance.

\noindent\textbf{Impact of Prompt Tuning.}
In Tab.~\ref{tab:6}, we compare DIFFender with prompt tuning against versions using "Empty prompt" and "Manual prompt" settings. For the manual prompt version, we set $prompt_L$ to "adversarial" and $prompt_R$ to "clean." The results show that prompt-tuned DIFFender achieves a substantial improvement in robust accuracy compared to the zero-shot versions, even with exposure to only a few attacked images. This underscores the effectiveness of prompt tuning.

\vspace{-0.1cm}

\subsection{Evaluation on ImageNet Classification}

\begin{table*}[t]
\centering
\small
\caption{\small Accuracy (\%) against attacks on ImageNet by Inception-v3 and Swin-S.}

  \scalebox{0.99}{
  \begin{tabular}{c|c|cc|cc|c|cc|cc}
  \toprule
               Models & \multicolumn{5}{c|}{Inception-v3}   & \multicolumn{5}{c}{Swin-S}         \\

\cmidrule(lr){1-11} 
&  & \multicolumn{2}{c|}{Adaptive} & \multicolumn{2}{c|}{Non-adaptive} &  & \multicolumn{2}{c|}{Adaptive} & \multicolumn{2}{c}{Non-adaptive} \\
\diagbox[height=0.6cm]{Defense }{Attack}       & Clean & AdvP & LaVAN & GDPA & RHDE & Clean & AdvP & LaVAN & GDPA & RHDE \\ \midrule
Undefended     & 100.0 & 0.0 & 8.2 & 64.8 & 39.8 & 100.0 & 1.6 & 3.5 & 78.1 & 51.6 
\\
JPEG~\cite{dziugaite2016study}              & 48.8 & 0.4 & 15.2 & 64.8 & 13.3 & 85.2 & 0.8 & 5.9 & 77.0 & 38.7 
\\
SS~\cite{xu2017feature}              & 72.7 & 1.2 & 14.8 & 57.8 & 16.4 & 86.3 & 2.3 & 5.5 & 68.8 & 34.8 
\\
DW~\cite{hayes2018visible}             & 87.1 & 1.2 & 9.4 & 62.5 & 28.5 & 88.3 & 0.0 & 5.1 & 77.3 & 66.0 
\\
LGS~\cite{naseer2019local}            & 87.9 & 55.5 & 50.4 & 67.2 & 49.6 & 89.8 & 65.6 & 59.8 & 82.0 & 69.1 
\\
FNC~\cite{yu2021defending}            & 91.0 & 61.3 & 64.8 & 66.4 & 46.5 & 91.8 & 6.3 & 7.4 & 77.0 & 63.7 
\\
DiffPure~\cite{nie2022diffusion}         & 65.2 & 10.5& 15.2& 67.6& 44.9& 74.6& 18.4& 26.2& 77.7& 62.3
\\
SAC~\cite{liu2022segment}           & \textbf{92.8} &84.2   &65.2   &68.0   &41.0   &93.6   &92.8   &84.6   &79.3   &54.9 
\\
Jedi~\cite{tarchoun2023jedi}           & 92.2 & 67.6 & 20.3 & 74.6 & 47.7 & 93.4 & 89.1 & 12.1 & 78.1 & 67.6 
\\

DIFFender & 91.4 & \textbf{88.3} & \textbf{71.9} & \textbf{75.0} & \textbf{53.5} & \textbf{93.8} & \textbf{94.5} & \textbf{85.9} & \textbf{82.4} & \textbf{70.3} \\

\bottomrule
\end{tabular}
 }
  \label{tab:1}
  \vspace{-0.2cm}
\end{table*}

\noindent\textbf{Quantitative Results.}
Tab.~\ref{tab:1} presents the experimental results, with the highest accuracy highlighted in bold. From these results, we can draw several key conclusions:

\begin{figure*}[h]
  \centering
  \includegraphics[width=0.95\linewidth]{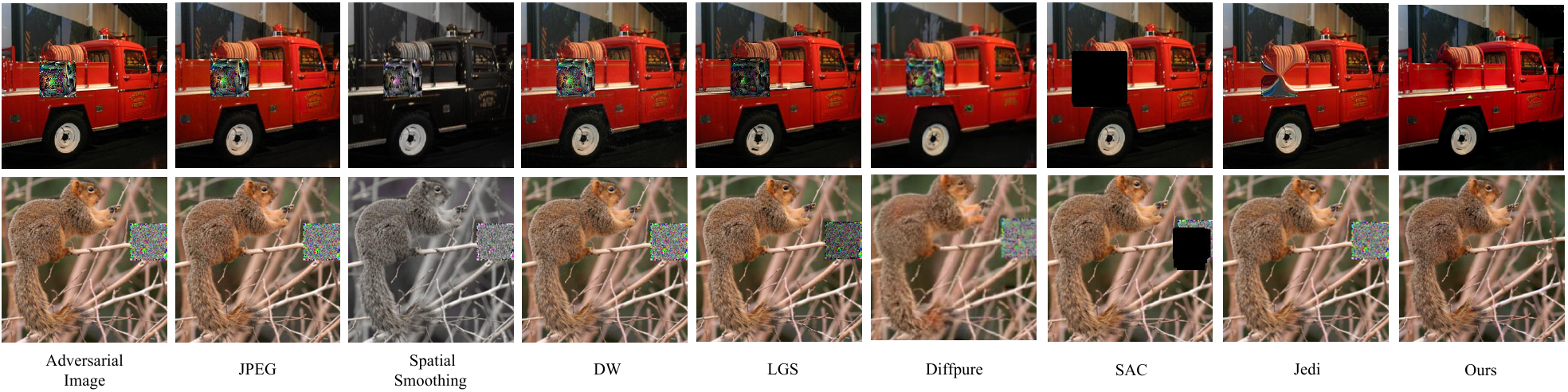}
  \caption{\small Visualization of ImageNet examples. DIFFender’s restored images display no residual traces of the adversarial patch, with notably detailed recovery (e.g., the restoration of tree branches in the second column).}
  \label{fig:3}
\end{figure*}

(1) DIFFender excels in defense effectiveness. Under adaptive attacks that exploit gradients, such as BPDA+AdvP and BPDA+LaVAN, DIFFender demonstrates exceptional performance, even with only an 8-shot tuning process. While some attacks, like GDPA, may not be as effective, DIFFender still achieves the highest robust accuracy. This is due to its foundation on a unified diffusion framework. Leveraging the AAP phenomenon, the diffusion model effectively identifies and removes adversarial regions while maintaining high-quality, diverse restoration that aligns closely with the distribution of clean data. Moreover, the inherent stochasticity of the diffusion model provides robust defense mechanisms \cite{he2019parametric}, making DIFFender a well-suited "defender" for adaptive attacks.

(2) Traditional image processing defenses show limitations. Methods like JPEG, SS, and DW experience a significant drop in robust accuracy under adaptive attacks, primarily because their gradients can be easily exploited. Although approaches such as LGS, FNC, SAC, and Jedi are designed with adaptive attack robustness—FNC, for instance, shows respectable robust accuracy on Inception-v3—their effectiveness diminishes on different architectures like Swin-S. This is likely because FNC’s feature norm clipping is specifically tailored for CNNs, whereas DIFFender’s generalization ability extends across different classifier types.

(3) DIFFender generalizes well to unseen attacks. In the experiments, DIFFender was tuned specifically for the AdvP method using 8-shot prompt tuning, yet it also performs well against other attacks. While Jedi exhibits strong robustness against certain attacks like AdvP, its robust accuracy drops significantly against others, such as LaVAN. This may be due to the autoencoder used by Jedi being trained under a specific style, limiting its generalization.

(4) Although RHDE is less threatening to undefended classifiers compared to adaptive meaningless attacks, it presents a greater challenge to defense methods due to its use of irregular, naturally-appearing patches. Nonetheless, DIFFender achieves the best defense results against RHDE without prior exposure to these patches. Additionally, DIFFender’s adaptability, facilitated by the prompt tuning module, allows for a few-shot tuning to further enhance performance against naturalistic patch attacks.

(5) While DiffPure performs well against global perturbations constrained by $\ell_p$-norms, it struggles with patch attacks. As shown in Tab.~\ref{tab:1}, when tested against AdvP and LaVAN, the Inception-v3 model purified by DiffPure only maintains robust accuracy rates of 10.5\% and 15.2\%, respectively, consistent with our observations in Sec.~\ref{sec:4-1}.

\noindent\textbf{Qualitative Results.}
Fig.~\ref{fig:3} illustrates the defense results against patch attacks. FNC, which suppresses feature maps during inference, is not shown in the figure. Other methods like JPEG and DW show only minor changes in reconstructed images and fail to defend against adaptive attacks. Images processed by Spatial Smoothing exhibit color distortion and remain vulnerable. LGS visibly suppresses the patch area, improving robust accuracy somewhat, but fails to completely eliminate the patch. Both Jedi and SAC encounter difficulties with localization in certain scenarios, as seen in the second row of Fig.~\ref{fig:3}, and Jedi’s restoration results are incomplete. In contrast, DIFFender’s restored images display no traces of the patch, with outstanding restoration details.

\subsection{Evaluation on Face Recognition}\label{sec:5-3}

\noindent\textbf{Experimental Settings.}
Face recognition presents a challenging task due to the rich diversity in facial expressions, combined with external factors such as lighting conditions and viewing angles. We conducted experiments on the LFW dataset \cite{huang2008labeled}, using two adversarial patch attacks: RHDE \cite{wei2022adversarial} and GDPA \cite{li2021generative}.

\begin{figure}[h]
  \centering
  \includegraphics[width=0.68\linewidth]{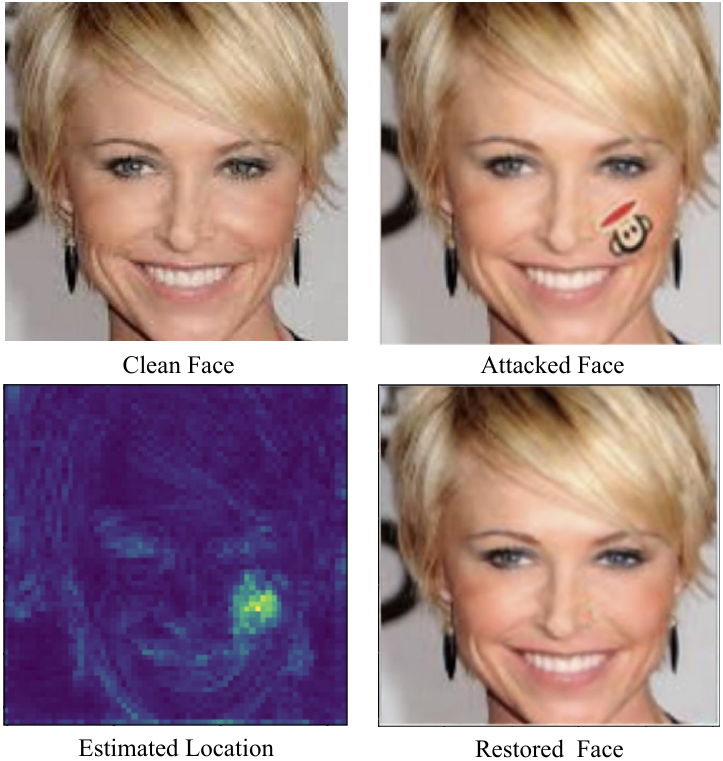}
  \caption{\small Visualization of LFW examples under natural RHDE attacks, with localization and restoration performed by DIFFender.}
  \label{fig:face}
\end{figure}

\begin{table}[h]
  \caption{\small Accuracy against patch attacks on LFW by FaceNet.}
    \centering
      \scalebox{1.2}{
    \centering
    \begin{tabular}{c|ccc}
    \toprule
                 & \multicolumn{3}{c}{FaceNet} \\ 
\textbf{Defense} & Clean    & GDPA    & RHDE  \\ \midrule
Undefended      & 100.0 & 56.3 & 42.8 
\\
JPEG~\cite{dziugaite2016study}             & 44.1 & 16.8 & 17.8 \\
SS~\cite{xu2017feature}                  & 19.9 & 8.2 & 3.5 \\
DW~\cite{hayes2018visible}               & 37.1 & 15.2 & 7.2 \\
LGS~\cite{naseer2019local}               & 60.9 & 71.9 & 53.5 \\
FNC~\cite{yu2021defending}                & 100.0 & 39.8 & 39.3 \\
SAC~\cite{liu2022segment}                 & 100.0 & 77.3 & 43.2 \\
Jedi~\cite{tarchoun2023jedi}               & 100.0 & 74.2 & 43.9 \\
DIFFender (EP)        & 100.0 & 79.3 & 57.2 \\
DIFFender (MP)         & 100.0 & 77.0 & 57.2 \\
DIFFender         & \textbf{100.0} & \textbf{81.1} & \textbf{60.7}  \\
    \bottomrule
  \end{tabular}}
  \label{tab:7}
\end{table}

\begin{figure*}[h]
  \includegraphics[width=0.8\linewidth]{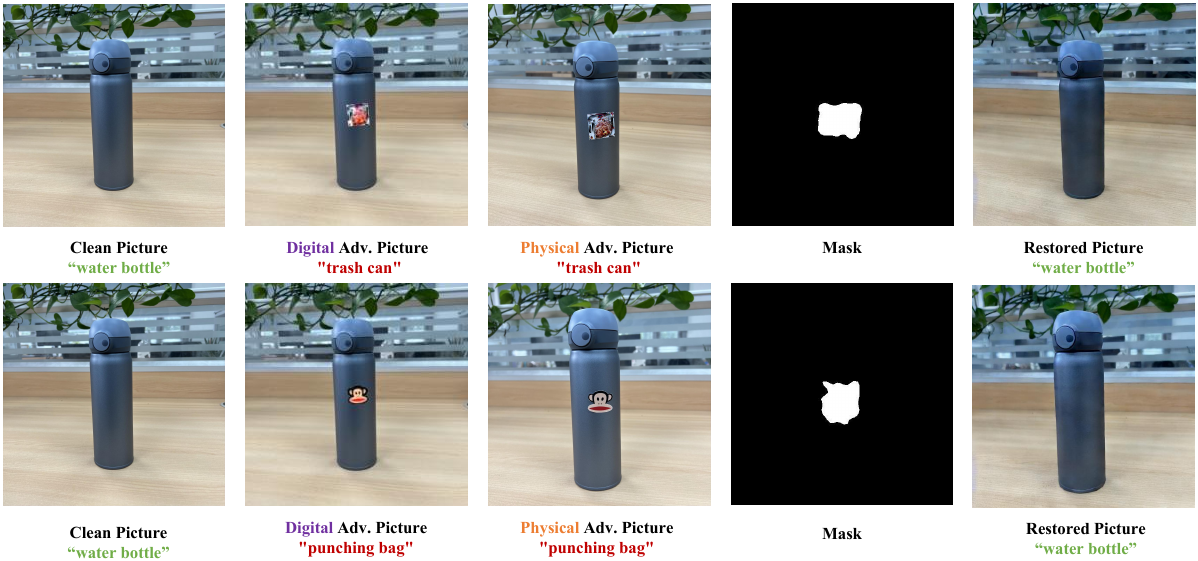}
  \centering
  \caption{\small 
    Demonstrations of DIFFender's defense in the physical world against both meaningless and natural patch attacks. The mask edges may slightly extend beyond the patch region, aiding in restoration and helping to maintain consistency in the restored image.
  }
  \label{fig:6}
\end{figure*}

\noindent\textbf{Experimental Results.}
The results on the LFW dataset are presented in Tab.~\ref{tab:7}. DIFFender achieves the highest robust accuracy under both the GDPA and RHDE attacks while maintaining high clean accuracy. Notably, DIFFender was not specifically re-tuned for facial recognition tasks, underscoring its generalizability across different scenarios and attack methods. In contrast, methods like JPEG, SS, and FNC show low robust accuracy. This is likely because facial recognition classifiers heavily rely on critical local features, and preprocessing the entire image can disrupt these important details. Fig.~\ref{fig:face} illustrates DIFFender’s effectiveness against attack, where it accurately identifies the location of adversarial patches and achieves excellent restoration.

\subsection{Evaluation in the Physical World}\label{sec:Physical_Extension}

We also conducted experiments in real-world settings, selecting 10 common object categories from ImageNet and performing two types of patch attacks (naturalistic and meaningless)~\cite{wei2022physically}. Our approach involved first generating digital-world attack results and then placing stickers on real-world objects in the same positions. We tested DIFFender under various conditions, including different angles (rotations) and distances. Qualitative results are shown in Fig.~\ref{fig:6}, while quantitative results are provided in Tab.~\ref{tab:8}, each configuration is based on 256 successfully classified frames from the selected objects. The results demonstrate that DIFFender maintains robust defensive capabilities across a range of physical alterations, proving its effectiveness in real-world scenarios.

\begin{table}[t]
\caption{\small Quantitative result of meaningless physical attacks on the Inception-v3 at different angles and distances. }
\centering
\vspace{0.2cm}
  \label{tab:8}
\centerline{
\scalebox{0.99}{
\begin{tabular}{c|ccccc}
 \toprule
               & 0°   & yaw ±15° & yaw ±30° & pitch ±15° & distance \\ \midrule
Undefended     & 28.9& 34.8& 41.8& 36.7& 35.9\\
Jedi \cite{tarchoun2023jedi}          & 61.7 & 57.8 & 66.4 & 63.3 & 62.1 \\
DIFFender & \textbf{80.9} & \textbf{76.6} & \textbf{77.7} & \textbf{75.4} & \textbf{73.8} \\ \bottomrule
\end{tabular} 
}
}
\vspace{0.2cm}
\end{table}

\section{Experiments in the Infrared Domain}\label{sec:Experiments_Infrared}

\subsection{Experimental Settings}
We use the LLVIP~\cite{jia2021llvip} dataset to conduct attacks and test defenses. Similar to Zhu et al.~\cite{
zhu2021fooling,zhu2022infrared}, we customize the parts of the LLVIP images that contain pedestrians. The test set contains 1220 images, and the training set contains 3784 images. As the final samples for the attack, we follow the infrared patch attack experimental setup from~\cite{wei2023infrared,wei2023unified} and select 128 images from the dataset that the target model can recognize with high probability. Therefore, the average precision (AP) in clean conditions is 100\%. In the subsequent experiments, we assume the size of the adversarial patch to be 150 pixels, which is about 15\% of the target object size. By default, the confidence threshold is set to 0.5, and the attack is considered successful when the detection confidence falls below 0.5. The evaluation metric we use is the Attack Success Rate (ASR), which represents the ratio of successfully attacked images out of all test images, to assess the performance of the defenses.

Following the experimental setup of~\cite{wei2023infrared,wei2023unified}, we evaluate the effectiveness of DIFFender in pedestrian detection by selecting representative mainstream detectors, specifically the one-stage detector YOLOv3~\cite{redmon2018yolov3} and the two-stage detector Faster RCNN~\cite{ren2015faster}. For each detector, we use the officially pre-trained weights as the initial weights and then retrain the model on the training dataset. These models are subsequently used as the target models in the attack and defense experiments. Additionally, we evaluate the latest infrared patch attack methods, AIP~\cite{wei2023infrared} and UAP~\cite{wei2023unified}, to test the defense performance of  DIFFender. Furthermore, we select five common defense methods for comparison: PixelMask~\cite{agarwal2021cognitive}, Bit Squeezing~\cite{xu2017feature}, JPEG Compression~\cite{dziugaite2016study}, Spatial Smoothing~\cite{xu2017feature}, and Total Variation Minimization~\cite{guo2017countering}. 

\begin{table*}[!t]
 \centering
\small
\caption{\small ASR (\%) of defenses against attacks on LLVIP by  YOLOv3 and  Faster RCNN.}
  \scalebox{0.99}{
  \begin{tabular}{c|ccc|ccc}
  \toprule
& \multicolumn{3}{c}{YOLOv3}                & \multicolumn{3}{c}{Faster RCNN}                \\
Defense                      & Clean        & AIP~\cite{wei2023infrared}           & UAP~\cite{wei2023unified}           & Clean        & AIP~\cite{wei2023infrared}              & UAP~\cite{wei2023unified}           \\
\midrule
Undefended                   & 0.0          & 84.4          & 89.1          & 0.0          & 86.7          & 78.1          \\
PixelMask~\cite{agarwal2021cognitive}                    & 0.8          & 82.0          & 78.1          & \textbf{0.0} & 85.2          & 78.1          \\
Bit squeezing~\cite{xu2017feature}                & \textbf{0.0} & 82.8          & 70.3          & 0.8          & 82.8          & 71.9          \\
JPEG compression~\cite{dziugaite2016study}             & 1.6          & 83.6          & 62.5          & 2.3          & 85.9          & 74.2          \\
Spatial smoothing~\cite{xu2017feature}            & 0.8          & 83.6          & 78.9          & 0.8          & 85.9          & 75.0          \\
Total variation mini~\cite{guo2017countering} & 10.2         & 73.4          & 50.0          & 6.3          & 47.7          & 47.7          \\
 DIFFender                & 7.0          & \textbf{14.8} & \textbf{10.2} & 5.5          & \textbf{17.2} & \textbf{19.5} \\
\bottomrule
\end{tabular}
}
  \label{tab:infrared}
\end{table*}

\begin{figure*}[!h]
  \centering
  \includegraphics[width=0.99\linewidth]{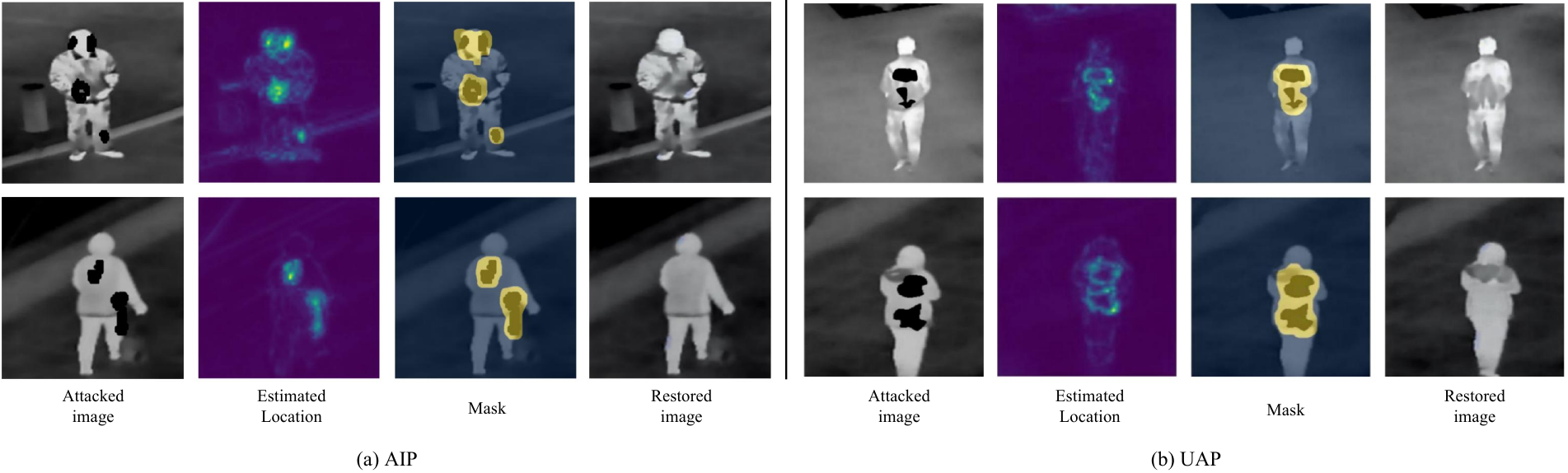} 
  \vspace{-1ex}\caption{\small Visualization with examples from LLVIP. (a) shows the defense results against AIP, and (b) shows the defense results against UAP.  DIFFender demonstrates accurate patch localization and restoration capabilities, achieving effective defense even in the presence of multiple irregular infrared patch attacks.}
  \label{fig:infrared}
\end{figure*}

\subsection{Ablation Studies}

\begin{table}[!t]
 \centering
\small
\caption{\small Ablation Study for the IDC token of  DIFFender, where No-IDC, L-IDC, R-IDC, and L\&R-IDC represent the scenarios of no IDC token, IDC token added to the localization/restoration/both stages, respectively.}
  \scalebox{0.99}{
  \begin{tabular}{c|ccc}
  \toprule
\multicolumn{1}{c}{}            & \multicolumn{3}{c}{YOLOv3}                                                    \\
\multicolumn{1}{c}{Defense}     & \multicolumn{1}{l}{Clean} & \multicolumn{1}{l}{AIP~\cite{wei2023infrared}} & \multicolumn{1}{l}{UAP~\cite{wei2023unified}} \\
\midrule
 DIFFender (No-IDC)   & 22.7                      & 27.3                    & 24.2                    \\
 DIFFender (L-IDC)    & 19.5                      & 20.3                    & 21.9                    \\
 DIFFender (R-IDC)    & 14.8                      & 18.0                    & 15.6                    \\
 DIFFender (L\&R-IDC) & \textbf{7.0}                   & \textbf{14.8}           & \textbf{10.2}          \\
\bottomrule
\end{tabular}
}
  \label{tab:IDC}
\end{table}

\begin{table}[h]
 \centering
\renewcommand\arraystretch{1.2}
\caption{\small Ablation study for new loss functions of DIFFender.}
  \scalebox{1.1}{
        \begin{tabular}{cc|ccc} 
            \toprule

                     &            & \multicolumn{3}{c}{YOLOv3}                                                    \\
$L_{IE}$ & $L_{TNC}$ & \multicolumn{1}{l}{Clean} & \multicolumn{1}{l}{AIP~\cite{wei2023infrared}} & \multicolumn{1}{l}{UAP~\cite{wei2023unified}} \\
\midrule
         &            & 25.0                      & 28.9                    & 24.2                    \\
\checkmark        &            & 18.0                      & 21.9                    & 18.0                    \\
         & \checkmark          & 4.7                       & 23.4                    & 21.9                    \\
\checkmark        & \checkmark          & \textbf{7.0}              & \textbf{14.8}           & \textbf{10.2}   \\
\bottomrule
          \end{tabular}
         }
          \label{tab:newloss}
\end{table}

\begin{table}[!t]
 \centering
\small
\caption{\small Comparison of original DIFFender with infrared-version DIFFender.}
  \scalebox{0.99}{
  \begin{tabular}{c|ccc}
  \toprule
\multicolumn{1}{c}{}        & \multicolumn{3}{c}{YOLOv3}                                                    \\
\multicolumn{1}{c}{Defense} & \multicolumn{1}{l}{Clean} & \multicolumn{1}{l}{AIP~\cite{wei2023infrared}} & 
\multicolumn{1}{l}{UAP~\cite{wei2023unified}} \\
\midrule
DIFFender (original)                   & 42.2                      & 47.7                    & 24.2                    \\
 DIFFender               &  \textbf{7.0}                       & \textbf{14.8}           & \textbf{10.2}         \\
\bottomrule
\end{tabular}
}
  \label{tab:comparison}
\end{table}

In Tables Tab.~\ref{tab:IDC} and Tab.~\ref{tab:newloss}, we conduct ablation experiments on the IDC token and the new loss functions, respectively, and in Tab.~\ref{tab:comparison}, we compare the results of  DIFFender with the original unmodified DIFFender on infrared data.

The results in Tab.~\ref{tab:IDC} demonstrate that the proposed IDC token provides benefits to both the localization and restoration modules of  DIFFender, with more significant gains in the restoration module. This may be because the domain-specific limiting effect of the IDC token can directly restrict the restoration module's output, preventing the generation of RGB patches that would contaminate the entire infrared image, thereby achieving more pronounced defense effects. As shown in the table, the  DIFFender with IDC tokens added to both the localization and restoration modules achieves the best defense results, confirming the effectiveness of the IDC token.

In Tab.~\ref{tab:newloss}, we perform ablation experiments on the newly introduced loss functions. We find that $L_{IE}$ primarily improves the performance of defense methods when facing attacks. This may be because edge information plays a crucial role in distinguishing infrared adversarial patches in the localization module. Additionally, $L_{TNC}$ not only enhances attack performance but also improves the performance on clean images. Ultimately,  DIFFender trained with both loss functions together achieves the best performance.

Finally, the comparison in Tab.~\ref{tab:comparison} shows that although the original unmodified DIFFender still has some effect on the infrared data defense task, demonstrating the potential of the DIFFender method to extend to other domain data, the performance of  DIFFender on infrared patches is significantly improved with the addition of the IDC token and new loss functions. This can be achieved with only a few-shot prompt-tuning, further demonstrating the scalability and applicability of  DIFFender in the infrared domain.

\subsection{Evaluation in Digital World}
The experimental results are shown in Tab.~\ref{tab:infrared}, and the visual results can be seen in Fig.~\ref{fig:infrared}. Most methods lost their defensive effectiveness against infrared patch attacks because previous defense methods were not designed for infrared patch attacks and overlooked the characteristics of infrared data. Compared to the previous defense methods,  DIFFender achieved the best results in the defense tasks within the infrared domain. For the one-stage detector,  DIFFender reduced the ASR by 69.6 and 78.9 against the AIP and UAP methods, respectively. For the two-stage detector,  DIFFender reduced the ASR by 69.5 and 58.6 against AIP and UAP, respectively. This verifies that DIFFender effectively mitigates the threat posed by infrared patch attacks and enhances the robustness of the detector. Besides, it shows that DIFFender can be easily adapted to new tasks through simple prompt-tuning, highlighting its scalability.

\subsection{Evaluation in Physical World}\label{sec:Experiments_Infrared_physical}

\begin{figure}[!h]
  \centering
  \includegraphics[width=0.99\linewidth]{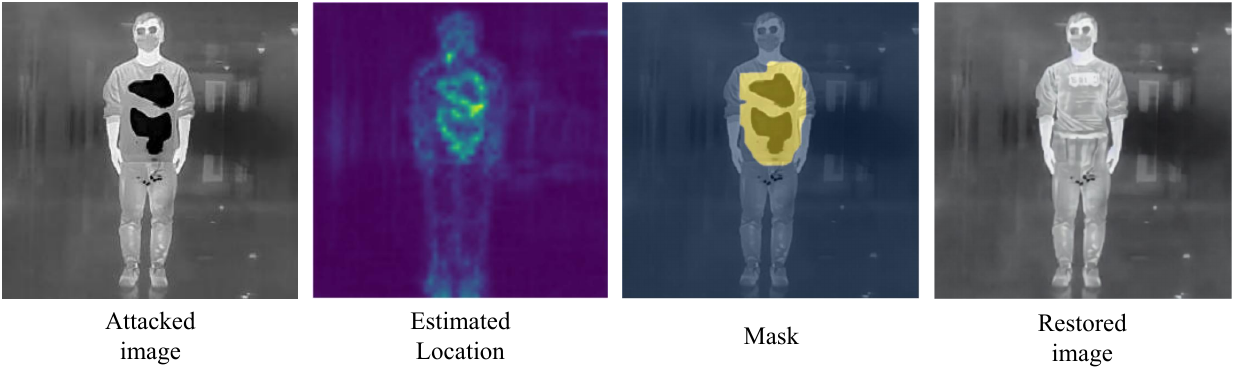} 
  \vspace{-1ex}\caption{\small Qualitative results of defense of UAP physical attacks on the YOLOv3 object detector.}
  \label{fig:infrared_phy}
\end{figure}

\begin{table}[!h]
 \centering
\small
\caption{\small Quantitative result of defense of UAP physical attacks on the YOLOv3 at different angles and distances.}
  \scalebox{0.99}{
  \begin{tabular}{c|ccccc}
  \toprule
          & 0°            & ±15°          & ±30°          & distance      & postures      \\
\midrule
UAP       & 76.6          & 65.6          & 53.9          & 69.5          & 63.3          \\
DIFFender & \textbf{19.5} & \textbf{24.2} & \textbf{28.9} & \textbf{18.8} & \textbf{21.1} \\
\bottomrule
\end{tabular}
}
  \label{tab:infrared_phy}
\end{table}

We further tested the defense results of  DIFFender in the physical world, where the attack experiments followed the setup of~\cite{wei2023infrared,wei2023unified}, including conducting physical adversarial attacks and subsequent defenses. The detailed process is similar to Sec.~\ref{sec:Physical_Extension}. We experimentally validated  DIFFender under various physical conditions, including 0-30 degree angles, different poses, and distances. Fig.~\ref{fig:infrared_phy} shows the qualitative demo of attack scenarios in the physical world. After defense processing by  DIFFender, the physical attacks with adversarial patches worn by the attacker failed. The quantitative results are shown in Tab.~\ref{tab:infrared_phy}. The experimental results demonstrate that  DIFFender can effectively defend against infrared patches even in real-world attack scenarios. It can be observed that the ASR of attack methods significantly decreased under various angles, distances, and poses, illustrating that our defense method maintains strong robustness even under various real-world conditions.

\section{Conclusion}
\label{sec:Conclusion}

We introduce \textbf{DIFFender}, a novel defense framework that harnesses a pre-trained unified diffusion model to address both the localization and restoration of patch attacks, building on the discovery of the Adversarial Anomaly Perception (AAP) phenomenon. To enhance the framework’s adaptability, we have developed a few-shot prompt-tuning algorithm that streamlines the tuning process, eliminating the need for extensive retraining. Our extensive experiments, covering image classification, face recognition, and real-world scenarios, demonstrate that DIFFender offers exceptional resilience even against adaptive attacks. Besides, it significantly enhances the generalization capabilities of pre-trained large models across various scenarios, classifiers, and attack methods, all while requiring only minimal prompt-tuning. Moreover, beyond its efficacy in the visible domain, DIFFender shows remarkable flexibility by seamlessly extending its defense capabilities to the infrared domain, offering a universal solution that can counter both infrared and visible adversarial patch attacks. This multi-modal applicability not only significantly reduces the success rate of patch attacks but also ensures the generation of realistic restored images, paving the way for broader applications of diffusion models and inspiring future research in this domain.







\ifCLASSOPTIONcaptionsoff
  \newpage
\fi



%

\bibliographystyle{splncs04}
\bibliography{egbib}

\begin{thebibliography}{10}
\providecommand{\url}[1]{\texttt{#1}}
\providecommand{\urlprefix}{URL }
\providecommand{\doi}[1]{https://doi.org/#1}

\bibitem{agarwal2021cognitive}
Agarwal, A., Vatsa, M., Singh, R., Ratha, N.: Cognitive data augmentation for adversarial defense via pixel masking. Pattern Recognition Letters  \textbf{146},  244--251 (2021)

\bibitem{athalye2018obfuscated}
Athalye, A., Carlini, N., Wagner, D.: Obfuscated gradients give a false sense of security: Circumventing defenses to adversarial examples. In: International conference on machine learning. pp. 274--283. PMLR (2018)

\bibitem{brown2017adversarial}
Brown, T.B., Man{\'e}, D., Roy, A., Abadi, M., Gilmer, J.: Adversarial patch. arXiv preprint arXiv:1712.09665  (2017)

\bibitem{carlini2017towards}
Carlini, N., Wagner, D.: Towards evaluating the robustness of neural networks. In: 2017 ieee symposium on security and privacy (sp). pp. 39--57. Ieee (2017)

\bibitem{chiang2020certified}
Chiang, P.y., Ni, R., Abdelkader, A., Zhu, C., Studer, C., Goldstein, T.: Certified defenses for adversarial patches. arXiv preprint arXiv:2003.06693  (2020)

\bibitem{deng2009imagenet}
Deng, J., Dong, W., Socher, R., Li, L.J., Li, K., Fei-Fei, L.: Imagenet: A large-scale hierarchical image database. In: 2009 IEEE conference on computer vision and pattern recognition. pp. 248--255. Ieee (2009)

\bibitem{dong2023benchmarking}
Dong, Y., Kang, C., Zhang, J., Zhu, Z., Wang, Y., Yang, X., Su, H., Wei, X., Zhu, J.: Benchmarking robustness of 3d object detection to common corruptions. In: Proceedings of the IEEE/CVF Conference on Computer Vision and Pattern Recognition. pp. 1022--1032 (2023)

\bibitem{dong2018boosting}
Dong, Y., Liao, F., Pang, T., Su, H., Zhu, J., Hu, X., Li, J.: Boosting adversarial attacks with momentum. In: Proceedings of the IEEE conference on computer vision and pattern recognition. pp. 9185--9193 (2018)

\bibitem{dziugaite2016study}
Dziugaite, G.K., Ghahramani, Z., Roy, D.M.: A study of the effect of jpg compression on adversarial images. arXiv preprint arXiv:1608.00853  (2016)

\bibitem{edwards2020study}
Edwards, D., Rawat, D.B.: Study of adversarial machine learning with infrared examples for surveillance applications. Electronics  \textbf{9}(8), ~1284 (2020)

\bibitem{goodfellow2014explaining}
Goodfellow, I.J., Shlens, J., Szegedy, C.: Explaining and harnessing adversarial examples. arXiv preprint arXiv:1412.6572  (2014)

\bibitem{gowal2019scalable}
Gowal, S., Dvijotham, K.D., Stanforth, R., Bunel, R., Qin, C., Uesato, J., Arandjelovic, R., Mann, T., Kohli, P.: Scalable verified training for provably robust image classification. In: Proceedings of the IEEE/CVF International Conference on Computer Vision. pp. 4842--4851 (2019)

\bibitem{guo2017countering}
Guo, C., Rana, M., Cisse, M., Van Der~Maaten, L.: Countering adversarial images using input transformations. arXiv preprint arXiv:1711.00117  (2017)

\bibitem{hayes2018visible}
Hayes, J.: On visible adversarial perturbations \& digital watermarking. In: Proceedings of the IEEE Conference on Computer Vision and Pattern Recognition Workshops. pp. 1597--1604 (2018)

\bibitem{he2019parametric}
He, Z., Rakin, A.S., Fan, D.: Parametric noise injection: Trainable randomness to improve deep neural network robustness against adversarial attack. In: Proceedings of the IEEE/CVF Conference on Computer Vision and Pattern Recognition. pp. 588--597 (2019)

\bibitem{ho2020denoising}
Ho, J., Jain, A., Abbeel, P.: Denoising diffusion probabilistic models. Advances in Neural Information Processing Systems  \textbf{33},  6840--6851 (2020)

\bibitem{huang2008labeled}
Huang, G.B., Mattar, M., Berg, T., Learned-Miller, E.: Labeled faces in the wild: A database forstudying face recognition in unconstrained environments. In: Workshop on faces in'Real-Life'Images: detection, alignment, and recognition (2008)

\bibitem{jia2021llvip}
Jia, X., Zhu, C., Li, M., Tang, W., Zhou, W.: Llvip: A visible-infrared paired dataset for low-light vision. In: Proceedings of the IEEE/CVF international conference on computer vision. pp. 3496--3504 (2021)

\bibitem{jing2021too}
Jing, P., Tang, Q., Du, Y., Xue, L., Luo, X., Wang, T., Nie, S., Wu, S.: Too good to be safe: Tricking lane detection in autonomous driving with crafted perturbations. In: Proceedings of USENIX Security Symposium (2021)

\bibitem{kang2023diffender}
Kang, C., Dong, Y., Wang, Z., Ruan, S., Su, H., Wei, X.: Diffender: Diffusion-based adversarial defense against patch attacks. European Conference on Computer Vision (ECCV)  (2024)

\bibitem{karmon2018lavan}
Karmon, D., Zoran, D., Goldberg, Y.: Lavan: Localized and visible adversarial noise. In: International Conference on Machine Learning. pp. 2507--2515. PMLR (2018)

\bibitem{li2021generative}
Li, X., Ji, S.: Generative dynamic patch attack. BMVC  (2021)

\bibitem{liao2018defense}
Liao, F., Liang, M., Dong, Y., Pang, T., Hu, X., Zhu, J.: Defense against adversarial attacks using high-level representation guided denoiser. In: Proceedings of the IEEE conference on computer vision and pattern recognition. pp. 1778--1787 (2018)

\bibitem{liu2022segment}
Liu, J., Levine, A., Lau, C.P., Chellappa, R., Feizi, S.: Segment and complete: Defending object detectors against adversarial patch attacks with robust patch detection. In: Proceedings of the IEEE/CVF Conference on Computer Vision and Pattern Recognition. pp. 14973--14982 (2022)

\bibitem{liu2021swin}
Liu, Z., Lin, Y., Cao, Y., Hu, H., Wei, Y., Zhang, Z., Lin, S., Guo, B.: Swin transformer: Hierarchical vision transformer using shifted windows. In: Proceedings of the IEEE/CVF international conference on computer vision. pp. 10012--10022 (2021)

\bibitem{madry2017towards}
Madry, A., Makelov, A., Schmidt, L., Tsipras, D., Vladu, A.: Towards deep learning models resistant to adversarial attacks. arXiv preprint arXiv:1706.06083  (2017)

\bibitem{moosavi2016deepfool}
Moosavi-Dezfooli, S.M., Fawzi, A., Frossard, P.: Deepfool: a simple and accurate method to fool deep neural networks. In: Proceedings of the IEEE conference on computer vision and pattern recognition. pp. 2574--2582 (2016)

\bibitem{naseer2019local}
Naseer, M., Khan, S., Porikli, F.: Local gradients smoothing: Defense against localized adversarial attacks. In: 2019 IEEE Winter Conference on Applications of Computer Vision (WACV). pp. 1300--1307. IEEE (2019)

\bibitem{nie2022diffusion}
Nie, W., Guo, B., Huang, Y., Xiao, C., Vahdat, A., Anandkumar, A.: Diffusion models for adversarial purification. arXiv preprint arXiv:2205.07460  (2022)

\bibitem{osahor2019deep}
Osahor, U.M., Nasrabadi, N.M.: Deep adversarial attack on target detection systems. In: Artificial intelligence and machine learning for multi-domain operations applications. vol. 11006, pp. 620--628. SPIE (2019)

\bibitem{rao2020adversarial}
Rao, S., Stutz, D., Schiele, B.: Adversarial training against location-optimized adversarial patches. In: Computer Vision--ECCV 2020 Workshops: Glasgow, UK, August 23--28, 2020, Proceedings, Part V 16. pp. 429--448. Springer (2020)

\bibitem{redmon2018yolov3}
Redmon, J., Farhadi, A.: Yolov3: An incremental improvement. arXiv preprint arXiv:1804.02767  (2018)

\bibitem{ren2015faster}
Ren, S., He, K., Girshick, R., Sun, J.: Faster r-cnn: Towards real-time object detection with region proposal networks. Advances in neural information processing systems  \textbf{28} (2015)

\bibitem{Rombach_2022_CVPR}
Rombach, R., Blattmann, A., Lorenz, D., Esser, P., Ommer, B.: High-resolution image synthesis with latent diffusion models. In: Proceedings of the IEEE/CVF Conference on Computer Vision and Pattern Recognition (CVPR). pp. 10684--10695 (June 2022)

\bibitem{Sharif2016Accessorize}
Sharif, M., Bhagavatula, S., Bauer, L., Reiter, M.K.: Accessorize to a crime: Real and stealthy attacks on state-of-the-art face recognition. In: Proceedings of the 2016 acm sigsac conference on computer and communications security. pp. 1528--1540 (2016)

\bibitem{sohl2015deep}
Sohl-Dickstein, J., Weiss, E., Maheswaranathan, N., Ganguli, S.: Deep unsupervised learning using nonequilibrium thermodynamics. In: International Conference on Machine Learning. pp. 2256--2265 (2015)

\bibitem{szegedy2016rethinking}
Szegedy, C., Vanhoucke, V., Ioffe, S., Shlens, J., Wojna, Z.: Rethinking the inception architecture for computer vision. In: Proceedings of the IEEE conference on computer vision and pattern recognition. pp. 2818--2826 (2016)

\bibitem{szegedy2013intriguing}
Szegedy, C., Zaremba, W., Sutskever, I., Bruna, J., Erhan, D., Goodfellow, I., Fergus, R.: Intriguing properties of neural networks. arXiv preprint arXiv:1312.6199  (2013)

\bibitem{tarchoun2023jedi}
Tarchoun, B., Ben~Khalifa, A., Mahjoub, M.A., Abu-Ghazaleh, N., Alouani, I.: Jedi: Entropy-based localization and removal of adversarial patches. In: Proceedings of the IEEE/CVF Conference on Computer Vision and Pattern Recognition. pp. 4087--4095 (2023)

\bibitem{wang2022guided}
Wang, J., Lyu, Z., Lin, D., Dai, B., Fu, H.: Guided diffusion model for adversarial purification. arXiv preprint arXiv:2205.14969  (2022)

\bibitem{wei2022adversarial}
Wei, X., Guo, Y., Yu, J.: Adversarial sticker: A stealthy attack method in the physical world. IEEE Transactions on Pattern Analysis and Machine Intelligence  \textbf{45}(3),  2711--2725 (2023)

\bibitem{wei2022simultaneously}
Wei, X., Guo, Y., Yu, J., Zhang, B.: Simultaneously optimizing perturbations and positions for black-box adversarial patch attacks. IEEE Transactions on Pattern Analysis and Machine Intelligence  \textbf{45}(7),  9041--9054 (2023)

\bibitem{wei2023unified}
Wei, X., Huang, Y., Sun, Y., Yu, J.: Unified adversarial patch for visible-infrared cross-modal attacks in the physical world. IEEE Transactions on Pattern Analysis and Machine Intelligence  (2023)

\bibitem{wei2022physically}
Wei, X., Pu, B., Lu, J., Wu, B.: Physically adversarial attacks and defenses in computer vision: A survey. arXiv preprint arXiv:2211.01671  (2022)

\bibitem{wei2023infrared}
Wei, X., Yu, J., Huang, Y.: Infrared adversarial patches with learnable shapes and locations in the physical world. International Journal of Computer Vision pp. 1--17 (2023)

\bibitem{wu2019defending}
Wu, T., Tong, L., Vorobeychik, Y.: Defending against physically realizable attacks on image classification. arXiv preprint arXiv:1909.09552  (2019)

\bibitem{xiao2022densepure}
Xiao, C., Chen, Z., Jin, K., Wang, J., Nie, W., Liu, M., Anandkumar, A., Li, B., Song, D.: Densepure: Understanding diffusion models for adversarial robustness. In: The Eleventh International Conference on Learning Representations (2022)

\bibitem{xiao2021improving}
Xiao, Z., Gao, X., Fu, C., Dong, Y., Gao, W., Zhang, X., Zhou, J., Zhu, J.: Improving transferability of adversarial patches on face recognition with generative models. In: Proceedings of the IEEE/CVF Conference on Computer Vision and Pattern Recognition. pp. 11845--11854 (2021)

\bibitem{xu2017feature}
Xu, W., Evans, D., Qi, Y.: Feature squeezing: Detecting adversarial examples in deep neural networks. arXiv preprint arXiv:1704.01155  (2017)

\bibitem{yin2019understanding}
Yin, P., Lyu, J., Zhang, S., Osher, S., Qi, Y., Xin, J.: Understanding straight-through estimator in training activation quantized neural nets. arXiv preprint arXiv:1903.05662  (2019)

\bibitem{yu2021defending}
Yu, C., Chen, J., Xue, Y., Liu, Y., Wan, W., Bao, J., Ma, H.: Defending against universal adversarial patches by clipping feature norms. In: Proceedings of the IEEE/CVF International Conference on Computer Vision. pp. 16434--16442 (2021)

\bibitem{zhang2018unreasonable}
Zhang, R., Isola, P., Efros, A.A., Shechtman, E., Wang, O.: The unreasonable effectiveness of deep features as a perceptual metric. In: Proceedings of the IEEE conference on computer vision and pattern recognition. pp. 586--595 (2018)

\bibitem{zhou2022learning}
Zhou, K., Yang, J., Loy, C.C., Liu, Z.: Learning to prompt for vision-language models. International Journal of Computer Vision  \textbf{130}(9),  2337--2348 (2022)

\bibitem{zhu2022infrared}
Zhu, X., Hu, Z., Huang, S., Li, J., Hu, X.: Infrared invisible clothing: Hiding from infrared detectors at multiple angles in real world. In: Proceedings of the IEEE/CVF Conference on Computer Vision and Pattern Recognition. pp. 13317--13326 (2022)

\bibitem{zhu2021fooling}
Zhu, X., Li, X., Li, J., Wang, Z., Hu, X.: Fooling thermal infrared pedestrian detectors in real world using small bulbs. In: Proceedings of the AAAI conference on artificial intelligence. vol.~35, pp. 3616--3624 (2021)

\end{thebibliography}

\end{document}